\documentclass[letterpaper]{article} % DO NOT CHANGE THIS
\usepackage[preprint]{aaai2027}  % DO NOT CHANGE THIS
\usepackage[hyphens]{url}  % DO NOT CHANGE THIS
\usepackage{graphicx} % DO NOT CHANGE THIS
\usepackage{natbib}  % DO NOT CHANGE THIS AND DO NOT ADD ANY OPTIONS TO IT
\usepackage{caption} % DO NOT CHANGE THIS AND DO NOT ADD ANY OPTIONS TO IT
\usepackage{algorithm}
\usepackage{algorithmic}

\usepackage{newfloat}
\usepackage{listings}
\DeclareCaptionStyle{ruled}{labelfont=normalfont,labelsep=colon,strut=off} % DO NOT CHANGE THIS
\floatstyle{ruled}
\newfloat{listing}{tb}{lst}{}
\floatname{listing}{Listing}

\usepackage{booktabs}

\title{Beyond the Mean: Multi-Moment Policy Optimization for LLM Reasoning}
\author{
Yijun Zhang\smash{\raisebox{0.5ex}{*}}
\hspace{1.0em}
Yule Xie\smash{\raisebox{0.5ex}{*}}
\hspace{1.0em}
Jiaxin Ding\smash{\raisebox{0.5ex}{\textdagger}}
\\
Xin Ding
\hspace{1.0em}
Fan Xu
\hspace{1.0em}
Haoxiang Zhang
\hspace{1.0em}
Luoyi Fu
}

\affiliations{
Shanghai Jiao Tong University
}

\usepackage{amsmath}
\usepackage{amssymb}
\newtheorem{proposition}{Proposition}
\newtheorem{theorem}{Theorem}
\usepackage{subcaption}
\usepackage{tabularx}
\usepackage{array}
\usepackage[table]{xcolor}
\usepackage{multirow}
\newcolumntype{Y}{>{\centering\arraybackslash}X}
\usepackage{mathtools}
\usepackage{extarrows}
\makeatletter
\newcommand{\blfootnote}[1]{%
  \begingroup
  \renewcommand{\thefootnote}{}%
  \renewcommand{\@makefntext}[1]{\noindent ##1}%
  \footnotetext{#1}%
  \endgroup
}
\makeatother

\begin{document}

\maketitle

\blfootnote{Preprint. \textsuperscript{*}Equal contribution. \textsuperscript{\textdagger}Corresponding author. Code is available at \url{https://github.com/e3trange/MMPO}.}

\enlargethispage{3\baselineskip}

\begin{abstract}
Reinforcement learning has become a central paradigm for improving the reasoning capabilities of large language models. Existing methods generally aim to reduce the failure probabilities induced across problems. In this paper, we introduce a moment-based perspective on policy optimization for LLM reasoning by treating the failure probability of a randomly sampled problem as a random variable and characterizing optimization objectives through its moments. Under this perspective, many existing methods optimize only a single moment of the failure-probability distribution, leaving its broader distributional structure largely uncharacterized. We propose \textbf{M}ulti-\textbf{M}oment \textbf{P}olicy \textbf{O}ptimization (MMPO), a novel policy optimization framework that jointly minimizes multiple moments of the failure-probability distribution. MMPO admits a direct operational interpretation as minimizing the expected truncated time required to obtain the first successful response. Beyond MMPO, we further develop a general moment-transformation framework that systematically induces different moment profiles and provides a unified view of a broader family of policy optimization objectives. Experiments across five mathematical reasoning benchmarks and models of different scales demonstrate that MMPO consistently outperforms strong baselines. We hope this moment-based perspective offers new insights into the design of policy optimization objectives for LLM reasoning.
\end{abstract}

\section{Introduction}
\label{sec:introduction}

\begin{figure}[t]
    \centering
    \includegraphics[width=\linewidth]{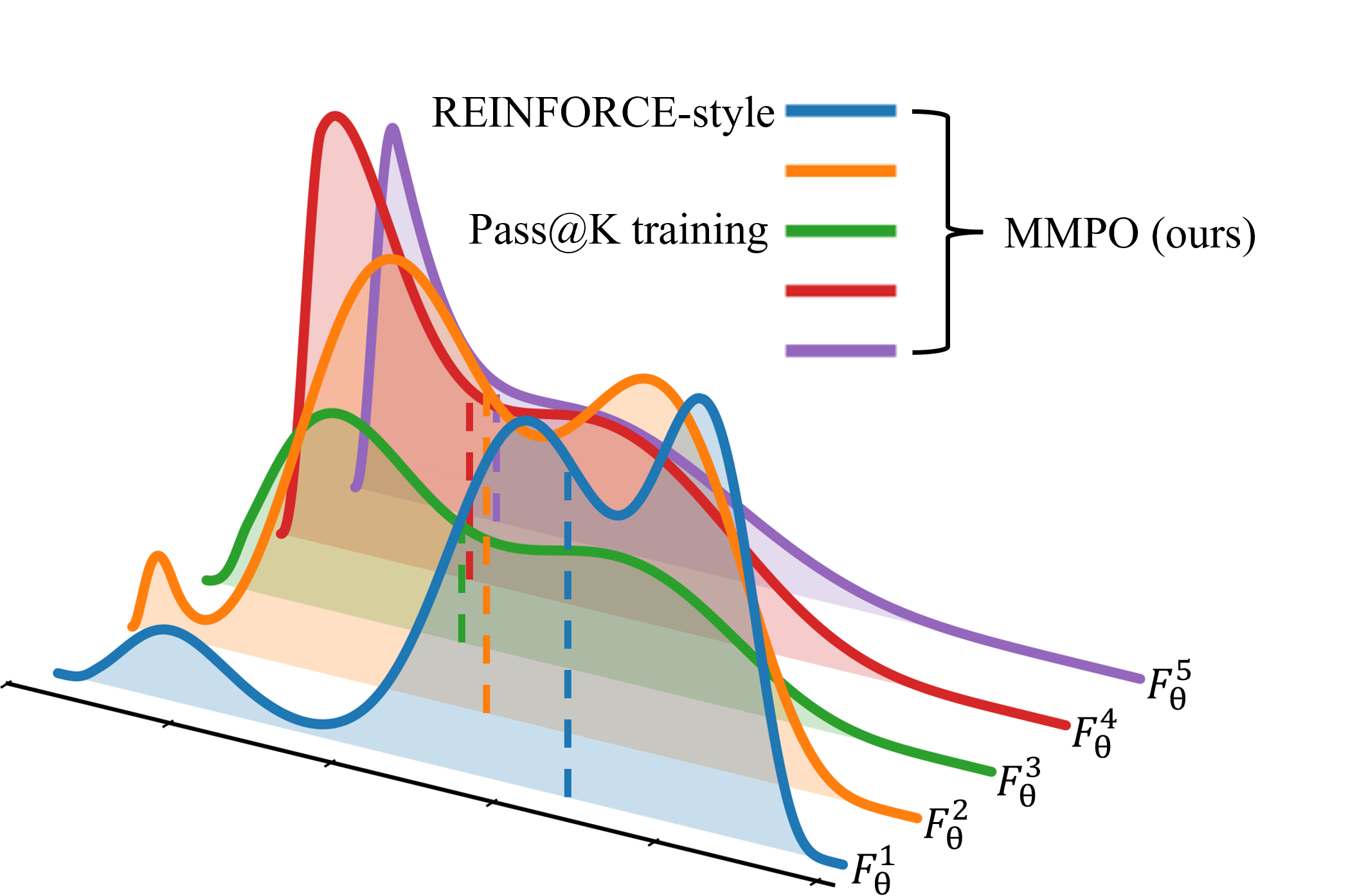}
    \caption{Illustration of single- and multi-moment optimization. $F_\theta$ denotes the failure-probability random variable across problems, and $F_\theta^i$ denotes the random variable obtained by raising $F_\theta$ to the $i$-th power. Each curve represents the probability density function of $F_\theta^i$, and the corresponding dashed line marks its mean $\mathbb{E}[F_\theta^i]$, i.e., the $i$-th moment of $F_\theta$. By jointly considering multiple moments, MMPO captures richer structural information.}
    \label{fig:moments_stack}
\end{figure}

Reinforcement Learning with Verifiable Rewards (RLVR) has emerged as a powerful paradigm for training large language models (LLMs)~\cite{lambert2024tulu}. By using rule-based verifiers, RLVR provides reliable outcome supervision for tasks such as search-based question answering~\cite{zhang2026information} and problem solving~\cite{guo2025deepseek}. In this work, we focus on mathematical reasoning, where each response sampled from the policy receives a binary reward indicating success or failure. Consequently, the policy induces a failure-probability distribution over problems. 

Existing RLVR objectives can be broadly expressed in terms of the failure-probability distribution. REINFORCE provides the standard expected-reward formulation~\cite{williams1992simple}, while GRPO~\cite{shao2024deepseekmath} and DAPO~\cite{yu2026dapo} introduce practical refinements for stable training. One line of work augments the objective to encourage exploration, including entropy-based regularization~\cite{jiang2025rethinking, zhang2026revisiting} and distribution-matching methods~\cite{zhu2025flowrl, li2026beyond}. Another line of work directly constructs optimization objectives from the failure distribution itself. Pass@$K$ training~\cite{chen2025pass, peng2025simko, walder2026pass} optimizes multi-rollout success toward solving the hard problems, whereas recent work MaxRL~\cite{tajwar2026maximum} derives a likelihood-oriented objective that connects reinforcement learning with maximum-likelihood learning. Despite their different motivations, these objectives lack a unified characterization of the distribution they optimize.

In this work, we introduce a moment-based perspective on policy optimization for LLM reasoning. \emph{To the best of our knowledge, our work is the first work to explicitly formulate policy optimization for LLM reasoning through the moments of the failure-probability distribution}. Under this perspective, REINFORCE-style methods optimize the first moment of the distribution, while pass@$K$ training optimizes a single higher-order moment. By the Hausdorff moment theorem~\cite{mnatsakanov2008hausdorff}, a distribution supported on $[0,1]$ is uniquely determined by its complete moment sequence. Optimizing only a single moment therefore captures only one aspect of the distribution, leaving its broader structure uncharacterized. Motivated by this observation, we propose Multi-Moment Policy Optimization (MMPO), which jointly minimizes multiple moments of the failure distribution (see Figure~\ref{fig:moments_stack}). MMPO admits a direct operational interpretation as minimizing the expected truncated number of rollout attempts required to obtain the first successful response, thereby providing a principled objective that balances average performance with greater attention to harder problems.

Building on the moment-based perspective, we further develop a generalized moment-transformation framework that unifies a broader family of policy optimization objectives, with different transformations inducing distinct optimization objectives. Under this framework, MaxRL can be interpreted as optimizing the moments of a transformed dristribution rather than those of the original one. We further prove that this family of objectives is strictly Schur-convex, revealing an explicit preference for more balanced success probabilities across problems. Experiments on five mathematical reasoning benchmarks and models of different scales demonstrate that MMPO consistently outperforms strong baselines. Our main contributions are summarized as follows:
\begin{itemize}
    \item We introduce a moment-based formulation of policy optimization for LLM reasoning and propose MMPO, which jointly optimizes multiple moments.
    
    \item We unify first-moment methods, pass@$K$ training, and MaxRL, and develop a moment-transformation framework with theoretical properties.
    
    \item We demonstrate across five mathematical reasoning benchmarks and models of different scales that MMPO consistently outperforms strong baselines.
\end{itemize}

\section{Related Work}
\label{sec:related_work}

\paragraph{Reinforcement Learning for LLM Reasoning.}
Reinforcement learning has become a standard approach for improving the reasoning capabilities of large language models. Given outcome-level supervision from rule-based verifiers, conventional methods optimize the expected reward through policy-gradient objectives. REINFORCE~\cite{williams1992simple} provides the basic formulation, while GRPO~\cite{shao2024deepseekmath} and DAPO~\cite{yu2026dapo} further introduce group-based advantage estimation and other strategies to improve training efficiency. A complementary line of work promotes exploration through entropy regularization~\cite{jiang2025rethinking, zhang2026revisiting}, distribution matching~\cite{zhu2025flowrl, li2026beyond}, diverse response generation~\cite{hu2025diversity}, or specialized rollout strategies~\cite{li2025treepo}. However, these methods remain primarily centered on first-moment optimization through the expected-reward objective.

\paragraph{Objectives beyond Expected Reward.}
Another line of work explores policy objectives beyond standard first-moment optimization. Pass@$K$ training~\cite{chen2025pass, peng2025simko, walder2026pass} maximizes the probability that at least one of $K$ sampled responses succeeds, while MaxRL~\cite{tajwar2026maximum} formulates RLVR from a maximum-likelihood perspective by maximizing the expected log success probability. Under our moment-based perspective, pass@$K$ training corresponds to optimizing a single higher-order moment, whereas MaxRL optimizes a weighted combination of multiple moments. In contrast, MMPO jointly minimizes multiple moments with uniform coefficients, providing a direct characterization of the failure-probability distribution beyond any single moment. Table~\ref{tab:objective_comparison} summarizes representative
policy objectives under the moment-based perspective. Here, $U_\lambda\sim\mathrm{Beta}(\lambda,1)$ is independent of
$F_\theta$, and MaxRL can be interpreted as optimizing multiple moments of a transformed random variable.

\section{Methodology}
\label{sec:methodology}

The methodology is organized as follows. Section~\ref{sec:method:preliminaries} establishes the basic problem formulation and revisits REINFORCE~\cite{williams1992simple} from a moment-based perspective, under which its optimization target can be interpreted as the expectation of a random variable. Section~\ref{sec:method:mmpo} then introduces MMPO, which is grounded in a practically meaningful objective of minimizing the expected time to first success, together with its objective construction and practical surrogate. Section~\ref{sec:method:theory} further provides a theoretical analysis of MMPO, clarifying its connections to pass@$K$ training methods~\cite{chen2025pass, peng2025simko, walder2026pass} and MaxRL~\cite{tajwar2026maximum}.

\subsection{Preliminaries}
\label{sec:method:preliminaries}

We consider a standard reinforcement learning formulation for LLM reasoning. Let $\mathcal{D}$ denote a probability distribution over problems $x\in\mathcal{X}$. For each problem $x \sim \mathcal{D}$, a policy model $\pi_\theta$ generates a response $y \sim \pi_\theta(\cdot \mid x)$. The generated response $y$ is evaluated by a verifier $r(x,y) \in \{0,1\}$, where $r(x,y)=1$ indicates a successful response and otherwise indicates failure. For any problem $x$, the success probability of a single rollout under the current policy is defined as
\begin{equation}
s_\theta(x)
=
\Pr_{y \sim \pi_\theta(\cdot \mid x)}\!\left[r(x,y)=1\right]
=
\mathbb{E}_{y \sim \pi_\theta(\cdot \mid x)}\!\left[r(x,y)\right].
\end{equation}

Correspondingly, we define the failure probability as $f_\theta(x)=1-s_\theta(x)$, which induces a random variable over problems. Let $X \sim \mathcal{D}$ denote the random problem instance, and define $F_\theta=f_\theta(X)$. The distribution of $F_\theta$ characterizes the heterogeneity of the current policy's failure probabilities across problems, indicating whether failures are concentrated on a few hard instances or broadly spread over the dataset.

REINFORCE optimizes the policy by maximizing the expected success probability of a single rollout over problems. Under the failure-probability view, this is equivalent to minimizing the expectation of $F_\theta$:
\begin{equation}
\label{eq:reinforce_objective}
\max_\theta \mathbb{E}_{X}\!\left[s_\theta(X)\right]
\Longleftrightarrow
\min_\theta \mathbb{E}_{X}\!\left[f_\theta(X)\right]
=
\min_\theta \mathbb{E}\!\left[F_\theta\right].
\end{equation}
The equality follows from the law of the unconscious statistician, $\int_{\mathcal{X}}f_\theta(x)\,\mathrm{d}\mathcal{D}(x)=\int_0^1z\,p_{F_\theta}(z)\,\mathrm{d}z$, where $p_{F_\theta}(z)$ is the density function induced by $F_\theta$ and describes the proportion of problems whose failure probability is near $z$. Objective~\eqref{eq:reinforce_objective} captures only the first moment of the random variable $F_\theta$, i.e., the average failure probability across problems. Recent methods such as GRPO~\cite{shao2024deepseekmath} and DAPO~\cite{yu2026dapo} introduce various refinements, yet their underlying objectives still remain centered on $\mathbb{E}\!\left[F_\theta\right]$. These methods do not explicitly characterize the distributional structure of $F_\theta$, leaving the behavior beyond the mean unmodeled. MMPO characterizes the failure distribution through the joint optimization of multiple moments, thereby extending beyond first-moment optimization.

\subsection{Multi-Moment Policy Optimization}
\label{sec:method:mmpo}

\begin{table}[t]
\centering
{
\small
\setlength{\tabcolsep}{2.5pt}
\begin{tabular}{lcc}
\toprule
Method & Objective & Moment View \\
\midrule
REINFORCE-style
& $\min_{\theta}\mathbb{E}[F_\theta]$
& single moment of $F_\theta$ \\

Pass@$K$ training
& $\min_{\theta}\mathbb{E}[F_\theta^{K}]$
& single moment of $F_\theta$ \\

MaxRL
& $\min_{\theta}\sum_{k=1}^{T}
   \mathbb{E}[F_\theta^{k}]/k$
& moments of $U_\lambda F_\theta$ \\

MMPO
& $\min_{\theta}\sum_{k=1}^{T}
   \mathbb{E}[F_\theta^{k}]$
& moments of $F_\theta$ \\
\bottomrule
\end{tabular}
}
\caption{Moment characterization and interpretation of representative policy objectives.}
\label{tab:objective_comparison}
\end{table}

\begin{algorithm}[t]
\caption{MMPO Training Workflow}
\label{alg:mmpo}
\begin{algorithmic}[1]
\REQUIRE Initial policy $\pi_{\theta}$, dataset $\mathcal{D}$, truncation order $T$, batch size $B$, group size $G$

\WHILE{not converged}
    \STATE Set $\pi_{\theta_{\mathrm{old}}}\leftarrow\pi_{\theta}$
    \STATE Sample a batch $\{x_i\}_{i=1}^{B}$ from $\mathcal{D}$
    \FOR{each problem $x_i$}
        \STATE Sample $\{y_{i,j}\}_{j=1}^{G}$ from $\pi_{\theta_{\mathrm{old}}}(\cdot\mid x_i)$
        \STATE Compute Advantage $\widehat{A}_{i,j}$ for each response $y_{i,j}$
    \ENDFOR
    \STATE Update $\theta$ by optimizing the clipped surrogate objective
\ENDWHILE

\RETURN $\pi_{\theta}$
\end{algorithmic}
\end{algorithm}

\paragraph{Population Objective.} MMPO extends the first-moment objective in~\eqref{eq:reinforce_objective} by jointly optimizing the first $T$ moments:
\begin{equation}
\label{eq:mmpo_objective}
\min_{\theta}\ \mathcal{J}_{T}(\theta),
\qquad
\mathcal{J}_{T}(\theta)
\coloneqq
\sum_{i=1}^{T}
\mathbb{E}\left[F_{\theta}^{i}\right].
\end{equation}
where $T$ denotes the truncation order. Interestingly,~\eqref{eq:mmpo_objective} admits a direct interpretation: it is equivalent to minimizing the expected truncated stopping time of the rollout process, namely, the expected number of attempts required to obtain the first successful response under a finite rollout budget.

To make such interpretation explicit, for each problem $x$, we define the first-success stopping time as
\begin{equation}
\begin{gathered}
S_{\theta}(x)
=
\inf\{t\geq 1:r(x,Y_t)=1\},\\
\text{where }(Y_t)_{t\geq 1}
\overset{\mathrm{i.i.d.}}{\sim}
\pi_{\theta}(\cdot\mid x).
\end{gathered}
\end{equation}
Since $S_{\theta}(x)$ follows a geometric distribution with success probability $s_{\theta}(x)$, i.e., $\Pr(S_{\theta}(x)=t)=f_{\theta}(x)^{t-1}s_{\theta}(x)$ for $t\geq 1$, its truncated expectation satisfies
\begin{equation}
\label{eq:practical_objective}
\mathbb{E}\!\left[
\min\{S_{\theta}(X),T+1\}
\,\middle|\,X=x
\right]
=
\sum_{i=0}^{T}f_{\theta}(x)^i.
\end{equation}
Thus, MMPO can be viewed as directly optimizing the time to first success, thereby exhibiting greater sensitivity to the long tail of difficult problems, which naturally require more rollout attempts before a successful response is obtained, than objectives based solely on average success probability.

\paragraph{Surrogate Objective.}
We next derive a practical surrogate for the population objective in~\eqref{eq:mmpo_objective}. Differentiating the objective with respect to $\theta$ gives
\begin{equation}
\label{eq:original_mmpo_gradient}
\nabla_{\theta}\mathcal{J}_{T}(\theta)
=
-
\mathbb{E}_{x\sim\mathcal{D}}
\left[
w_{T,\theta}(x)\nabla_{\theta}s_{\theta}(x)
\right],
\end{equation}
where
\begin{equation}
\label{eq:mmpo_weight}
w_{T,\theta}(x)
=
\sum_{k=1}^{T}
k f_{\theta}(x)^{k-1}
\end{equation}
is a problem-level weight induced by the joint optimization of the first $T$ moments. Using the score-function identity,
\begin{equation}
\nabla_{\theta}s_{\theta}(x)
=
\mathbb{E}_{y\sim\pi_{\theta}(\cdot\mid x)}
\left[
r(x,y)\nabla_{\theta}\log\pi_{\theta}(y\mid x)
\right],
\end{equation}
and subtracting the baseline, which leaves the expectation unchanged, the descent direction can be written as
\begin{equation}
\label{eq:mmpo_detailed_objective}
\begin{aligned}
-\nabla_{\theta}\mathcal{J}_{T}(\theta)
=
\mathbb{E}_{\substack{x\sim\mathcal{D}\\
y\sim\pi_{\theta}(\cdot\mid x)}}
\Big[
&w_{T,\theta}(x)
\left(r(x,y)-s_{\theta}(x)\right)\\
&\cdot\nabla_{\theta}\log\pi_{\theta}(y\mid x)
\Big].
\end{aligned}
\end{equation}

In practice, for each problem $x_i$ within a batch, we sample $G$ responses
$\{y_{i,j}\}_{j=1}^{G}$ and estimate~\eqref{eq:mmpo_detailed_objective}. A simple approach is to estimate both $s_{\theta}(x)$ and $w_{T,\theta}(x)$ from the sampled group. Specifically, for each $x_i$, we compute $
\widehat{s}_{i}
=
\frac{1}{G}\sum_{j=1}^{G}r(x_i,y_{i,j})
$ and $
\widehat{w}_{i}
=
\sum_{k=1}^{T}k(1-\widehat{s}_{i})^{k-1}$ by Eq.~\eqref{eq:mmpo_weight}. The corresponding advantage is given by
\begin{equation}
\label{eq:biased_advantage}
\widehat{A}_{i,j}^{\text{bias}}
=
\widehat{w}_{i}
\left(r(x_i,y_{i,j})-\widehat{s}_{i}\right).
\end{equation}
Although this plug-in estimator is generally biased, we find it empirically effective. Alternatively, when $T\leq G$, an unbiased estimator of the policy gradient direction can be constructed. Let $
M_{i,-j}
=
\sum_{\ell\neq j}\left(1-r(x_i,y_{i,\ell})\right)
$ denote the number of failures excluding the $j$-th rollout. An unbiased estimator of the policy gradient direction can be constructed using the following leave-one-out advantage coefficient:
\begin{equation}
\label{eq:unbiased_advantage}
\widehat{A}_{i,j}^{\mathrm{unb}}
=
\left[
\sum_{k=1}^{T}
k
\frac{\binom{M_{i,-j}}{k-1}}
{\binom{G-1}{k-1}}
\right]
\left(
r(x_i,y_{i,j})
-
1
+
\frac{M_{i,-j}}{G-1}
\right).
\end{equation}
Eq.~\eqref{eq:unbiased_advantage} further admits a particularly simple form when $T=G$. Let $N_i=\sum_{j=1}^{G}r(x_i,y_{i,j})$ denote the number of successful rollouts. Then we have
\begin{equation}
\label{eq:mmpo_simple_case}
\widehat{A}_{i,j}^{\mathrm{unb}}
=
\begin{cases}
\displaystyle
\frac{G(G+1)(G-N_i)}
{(G-1)N_i(N_i+1)},
& r(x_i,y_{i,j})=1,\\[10pt]
\displaystyle
-\frac{G(G+1)N_i}
{(G-1)(N_i+1)(N_i+2)},
& r(x_i,y_{i,j})=0.
\end{cases}
\end{equation}

Following PPO-style optimization~\cite{schulman2017proximal}, we maximize the corresponding clipped surrogate objective
\begin{equation}
\label{eq:mmpo_surrogate}
\begin{aligned}
\mathcal{L}_{\mathrm{MMPO}}(\theta)
=
\frac{1}{BG}
&\sum_{i=1}^{B}\sum_{j=1}^{G}
\min\Big(
\rho_{i,j}(\theta)\widehat{A}_{i,j},\\
&
\operatorname{clip}\!\left(
\rho_{i,j}(\theta),
1-\epsilon,
1+\epsilon
\right)\widehat{A}_{i,j}
\Big),
\end{aligned}
\end{equation}
where
$\rho_{i,j}(\theta)
=
\frac{\pi_{\theta}(y_{i,j}\mid x_i)}
{\pi_{\theta_{\mathrm{old}}}(y_{i,j}\mid x_i)}$
is the importance ratio, $\widehat{A}_{i,j}$ is the realized advantage, $\epsilon$ is the clipping parameter, and $B$ is the batch size. The derivation of the unbiased estimater is provided in Appendix~\ref{app:unbiased_estimator}, and the overall MMPO training procedure is summarized in Algorithm~\ref{alg:mmpo}.

\subsection{Theoretical Analysis}
\label{sec:method:theory}

\paragraph{Multi-Moment View.}
We first place REINFORCE-style methods, pass@K training methods, and MMPO within a unified moment-based framework. Let $\mu_k(\theta)=\mathbb{E}[F_\theta^k]$ denote the $k$-th raw moment of the failure-probability random variable $F_\theta$. As established in~\eqref{eq:reinforce_objective}, REINFORCE-style methods optimize the first moment $\mu_1(\theta)$. In comparison, maximizing the pass@$K$ objective, $\operatorname{Pass@}K(\theta)=1-\mu_K(\theta)$, is equivalent to minimizing the $K$-th raw moment $\mu_K(\theta)$. MMPO instead minimizes
$\sum_{k=1}^{T}\mu_k(\theta)$
in~\eqref{eq:mmpo_objective}, thereby jointly optimizing a moment sequence rather than only one of its coordinates. This formulation further provides a distributional perspective on policy optimization. Define the moment profile of $F_\theta$ as
$\mu(F_\theta)
=
\bigl(
\mu_1(\theta),
\mu_2(\theta),
\ldots
\bigr)$. Since $F_\theta$ is supported on the compact interval $[0,1]$, the classical Hausdorff moment problem implies that its complete moment sequence uniquely determines its probability distribution~\cite{mnatsakanov2008hausdorff,liu2016generating}. The moment profile $\mu(F_\theta)$ therefore provides a complete representation of how failure probabilities are distributed across problems.

Although a finite set of moments does not in general uniquely determine the distribution, \textbf{we argue that jointly and equally optimizing multiple moments is advantageous}, as it captures complementary aspects of the distribution that are more difficult to recover from any single moment alone. Notably, existing pass@$K$ methods improve final performance by annealing $K$~\cite{chen2025pass}. Under our moment-based view, such schedules admit a new interpretation: varying $K$ sequentially changes the optimized coordinate of the moment sequence. The empirical effectiveness of these schedules is therefore consistent with the view that different moments provide complementary optimization signals. In contrast to annealing among pass@$K$ objectives, MMPO provides a principled multi-moment formulation, as its combination of moments arises directly from the practically meaningful objective in Eq.\eqref{eq:practical_objective}. Moreover, whereas larger budgets $G$ in REINFORCE-style methods primarily reduce variance, \textbf{we propose leveraging a larger $G$ to characterize $F_\theta$ through a broader range of moments}.

The role of multiple moments can be further understood through tail-probability control. For any integer $\ell\geq 1$ and threshold $\tau\in(0,1]$, Markov's inequality gives
\begin{equation}
\Pr\left(F_\theta\geq\tau\right)
=
\Pr\left(F_\theta^\ell\geq\tau^\ell\right)
\leq
\frac{\mu_\ell(\theta)}{\tau^\ell}.
\end{equation}
Thus, each moment provides a distinct upper bound on $\Pr\left(F_\theta\geq\tau\right)$. Moreover, a reduction in one moment does not necessarily imply a reduction in another, since different moments may induce different orderings over policies. A concrete toy example is provided in Appendix~\ref{app:toy_example}.

\paragraph{Generalized Moment Transformation.}
A closely related recent work is MaxRL, which applies a Maclaurin expansion to the log-likelihood and truncates the resulting series at order $T$. Under our moment-based view, its objective can be equivalently written as $\mathrm{MaxRL}_T(\theta)=\sum_{k=1}^{T}\frac{1}{k}\mathbb{E}[F_\theta^k]$. The key distinction lies in the coefficients assigned to different moments: MMPO assigns a uniform coefficient of $1$ to each moment, whereas MaxRL adopts the harmonically decaying coefficient $1/k$. Consequently, the contribution of $\mathbb{E}[F_\theta^k]$ is progressively attenuated as $k$ increases in MaxRL, while MMPO preserves the influence of higher-order moments without such coefficient-level decay. We therefore argue that MMPO directly performs joint optimization over the moments of $F_\theta$, whereas MaxRL optimizes the moments of \textbf{another random variable}. The Interpretation is inspired by the classical Hausdorff moment
characterization~\cite{berg2005some}, summarized below.

\begin{theorem}[Hausdorff theorem]
\label{theorem:hausdorff_theorem}
Let $A=\{a_k\}_{k=0}^{\infty}$ be a real sequence with $a_0=1$, and define the forward-difference operator by $\Delta a_k=a_{k+1}-a_k$. Then $A$ is the moment profile of a random variable supported on $[0,1]$ if and only if
\begin{equation}
\label{eq:hausdorff_theorem_condition}
(-1)^n\Delta^n a_k \geq 0,
\qquad \forall\, n,k\geq 0.
\end{equation}
Moreover, whenever~\eqref{eq:hausdorff_theorem_condition} holds, there exists a unique $[0,1]$-valued random variable $U$ in distribution, such that
\begin{equation}
\label{eq:hausdorff_theorem_coe}
a_k=\mathbb{E}[U^k],
\qquad \forall\, k\geq 0.
\end{equation}
Furthermore, if $A$ and $C=\{c_k\}_{k=0}^{\infty}$ are both moment profiles, then $\{a_kc_k\}_{k=0}^{\infty}$ is also a moment profile.
\end{theorem}

\begin{figure}[t]
    \centering
    \begin{subfigure}[t]{0.40\linewidth}
        \centering
        \includegraphics[width=\linewidth]{\detokenize{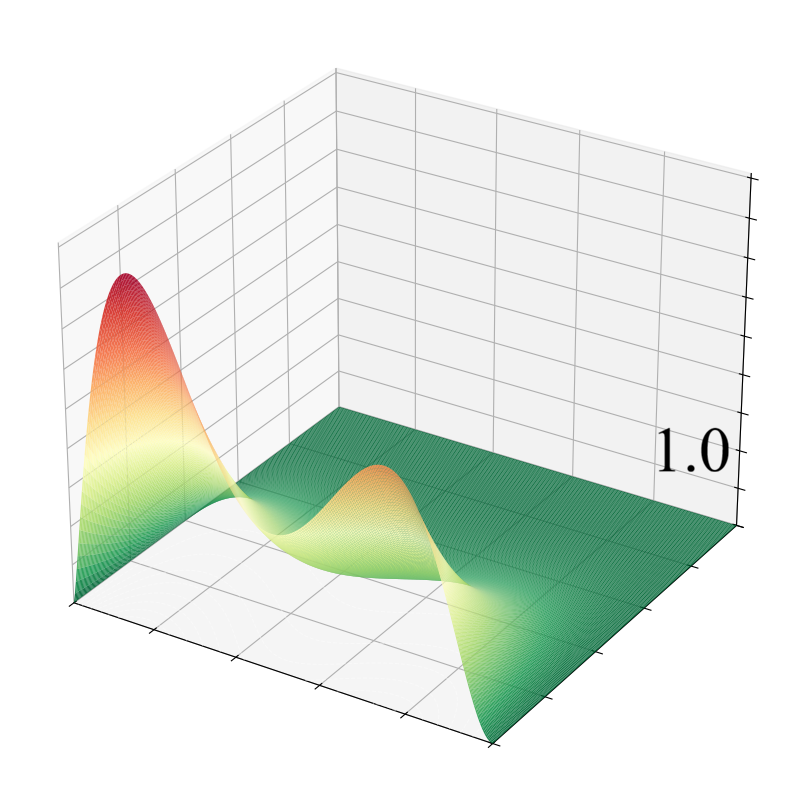}}
        \caption{$F_{\theta_1}$, $U\sim\mathrm{Beta}(1,2)$}
        \label{fig:f1-beta12}
    \end{subfigure}
    \begin{subfigure}[t]{0.40\linewidth}
        \centering
        \includegraphics[width=\linewidth]{\detokenize{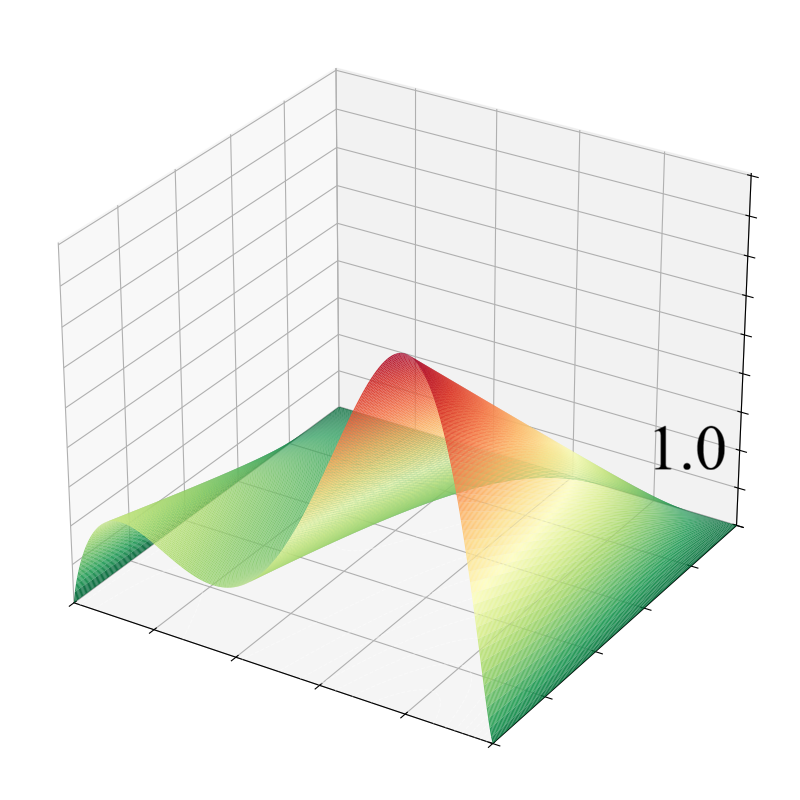}}
        \caption{$F_{\theta_2}$, $U\sim\mathrm{Beta}(1,2)$}
        \label{fig:f2-beta12}
    \end{subfigure}
    \begin{subfigure}[t]{0.40\linewidth}
        \centering
        \includegraphics[width=\linewidth]{\detokenize{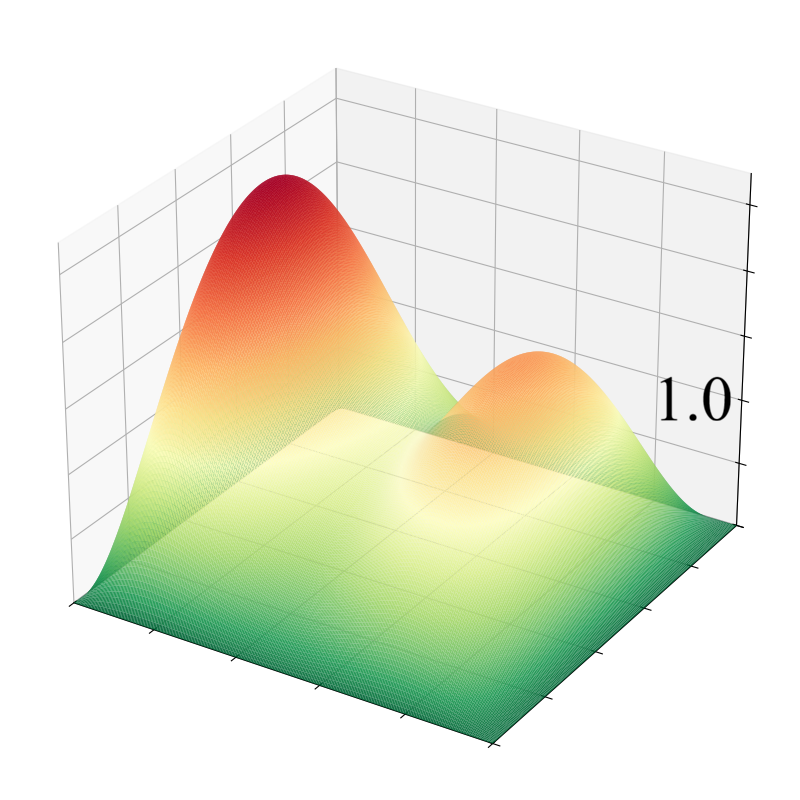}}
        \caption{$F_{\theta_1}$, $U\sim\mathrm{Beta}(2,2)$}
        \label{fig:f1-beta13}
    \end{subfigure}
    \begin{subfigure}[t]{0.40\linewidth}
        \centering
        \includegraphics[width=\linewidth]{\detokenize{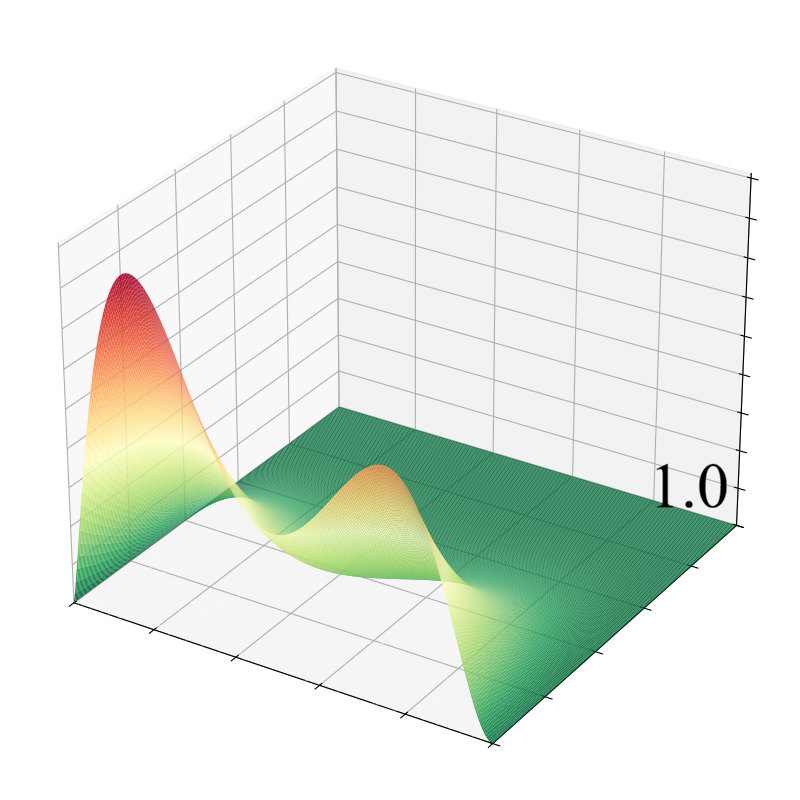}}
        \caption{$F_{\theta_1}$, $U\sim\mathrm{Beta}(1,5)$}
        \label{fig:f1-beta15}
    \end{subfigure}

    \caption{Joint density surfaces induced by different failure distributions $F_\theta$ and transformations $U$. The $x$- and $y$-axes represent realizations $f$ of $F_\theta$ and $u$ of $U$, respectively. For $f,u\in[0,1]$, the surface height is $z=p_{F_\theta}(f)p_U(u)$, where $p_{F_\theta}$ and $p_U$ denote their respective densities.}
    \label{fig:joint-density-surfaces}
\end{figure}

Constructively, we consider a $[0,1]$-valued
random variable $U_\lambda\sim\operatorname{Beta}(\lambda,1)$ for
$\lambda>0$, independent of $F_\theta$. Its $k$-th moment is given by
\begin{equation}
\mathbb{E}[U_\lambda^k]
=
\lambda\int_0^1 u^{k+\lambda-1}\,\mathrm{d}u
=
\frac{\lambda}{\lambda+k}.
\end{equation}
Since $\{\mathbb{E}[U_\lambda^k]\}_{k=0}^{\infty}$ is a moment profiles (satisfying ~\eqref{eq:hausdorff_theorem_condition}), it shows that $\{\mathbb{E}[(U_\lambda F_\theta)^k]\}_{k=0}^{\infty}$
is also a moment profile. Thus, MaxRL can be interpreted as optimizing the moments of the random variable $U_\lambda F_\theta$, rather than those of $F_\theta$ itself:
\begin{equation}
\begin{aligned}
&\sum_{k=1}^{T}\mathbb{E}\!\left[(U_\lambda F_\theta)^k\right]
=
\lambda\sum_{k=1}^{T}\frac{1}{\lambda+k}\mathbb{E}[F_\theta^k] \\
&\xrightarrow[\lambda\to0^+]{\text{up to positive scaling}}
\sum_{k=1}^{T}\frac{1}{k}\mathbb{E}[F_\theta^k]
=
\text{MaxRL}_T(\theta).
\end{aligned}
\end{equation}
Here, the outer factor $\lambda$ only rescales the gradient, while the relative coefficient between the first and the $k$-th moments approaches $k$, attenuating higher-order moments.

By Theorem~\ref{theorem:hausdorff_theorem}, we can naturally extend this construction to any $[0,1]$-valued random variable $U$ independent of $F_\theta$. Optimizing the moments of $UF_\theta$ induces a systematic reweighting of the moments of $F_\theta$. Furthermore, the transformation by $U$ contracts discrepancies between failure distributions, since $W_p\!\left(UF_{\theta_1},UF_{\theta_2}\right)\leq \left(\mathbb{E}[U^p]\right)^{1/p}W_p\!\left(F_{\theta_1},F_{\theta_2}\right)$, where $W_p$ denotes the $p$-th Wasserstein distance between the corresponding distributions. As shown in Figure~\ref{fig:joint-density-surfaces}, different choices of $F_\theta$ and $U$ induce markedly different joint density structures. The degenerate choice $U\equiv1$ directly characterizes $F_\theta$ and recovers MMPO, whereas MaxRL characterizes $U_\lambda F_\theta$ with a small $\lambda$. More generally, different choices of $U$ induce distinct optimization problems, which may be better suited to different reasoning settings or datasets.

\paragraph{Schur-Convexity.} We further establish a structural property shared by this family of generalized moment objectives, i.e., objectives of the form $\min_{\theta} \sum_{k=1}^{T}\mathbb{E}\left[(UF_\theta)^k\right]$ for any $[0,1]$-valued random variable $U$ independent of $F_\theta$. We show that all such objectives induce an explicit preference for more balanced failure probabilities across problems:
\begin{theorem}[Schur-Convex Moment Objectives]
\label{theorem:schur_convex_theorem}
Let $\mathbf{f}=(f_1,\ldots,f_n)\in[0,1]^n$ denote the failure probabilities of $n$ problems. For any $[0,1]$-valued random variable $U$ satisfying $\Pr(U>0)>0$ and any truncation order $T\geq2$, define
\begin{equation}
J_{U,T}(\mathbf{f})=\frac{1}{n}\sum_{i=1}^{n}\sum_{k=1}^{T}\mathbb{E}[U^k]f_i^k.
\end{equation}
Then $J_{U,T}$ is strictly Schur-convex in $\mathbf{f}$. Specifically, for any $\mathbf{g}=(g_1,\ldots,g_n)\in[0,1]^n$,
$
\mathbf{f}\succ\mathbf{g}
\Longrightarrow
J_{U,T}(\mathbf{f})>J_{U,T}(\mathbf{g}),
$
where $\mathbf{f}\succ\mathbf{g}$ means that $\mathbf{f}$ majorizes $\mathbf{g}$ (Appendix~\ref{app:schur_convexity}).
\end{theorem}

Here, $\mathbf{f}\succ\mathbf{g}$ indicates that $\mathbf{f}$ and $\mathbf{g}$ have the same mean, while $\mathbf{f}$ is more dispersed across problems. Accordingly, Theorem~\ref{theorem:schur_convex_theorem} establishes that the generalized moment objective exhibits an explicit preference for more balanced success probabilities across problems. Therefore, our objective encourages accuracy gains to be distributed more evenly across problems, rather than concentrated on easily solvable ones, thereby mitigating the tendency of GRPO to optimize primarily over solvable problems~\cite{qu2026pope} and promoting broader improvement across the problem distribution.

\paragraph{Controlled Reweighting toward Harder Problems.} As shown in Eq.~\eqref{eq:mmpo_weight}, implementing our multi-moment objective requires only an additional problem-level reweighting compared with GRPO. This reweighting assigns greater emphasis to problems with lower success probabilities, thereby directing optimization toward harder problems and encouraging broader exploration. However, aggressively emphasizing difficult problems, as in Pass@$K$ training, may come at the cost of reduced average performance. The following property shows that our reweighting alleviates this limitation:
\begin{proposition}[Moderate Reweighting]
\label{prop:reweight} For any two problems $x_{\mathrm{hard}}$ and
$x_{\mathrm{easy}}$ satisfying
$\rho=\frac{f_\theta(x_{\mathrm{hard}})}
{f_\theta(x_{\mathrm{easy}})}>1$, let
\begin{equation}
c_T(\rho)
=
\frac{
2\left(\rho^{T+1}-(T+1)\rho+T\right)
}{
T(T+1)(\rho-1)^2
}.
\end{equation}
Then $c_T(\rho)\geq1$, and the relative weights satisfy
\begin{equation}
\label{eq:weight_ratio_cmp}
1\leq
\frac{w_{T,\theta}(x_{\mathrm{hard}})}
     {w_{T,\theta}(x_{\mathrm{easy}})}
\leq
\frac{1}{c_T(\rho)}
\frac{W_{T,\theta}(x_{\mathrm{hard}})}
     {W_{T,\theta}(x_{\mathrm{easy}})},
\end{equation}
where
$
W_{T,\theta}(x)
=
Tf_\theta(x)^{T-1}
$
denotes the problem-level weight induced by the
pass@$T$ objective. In particular, $c_T(\rho)>2$ if and only if
\begin{equation}
\sum_{k=1}^{T-1}(T-k)\rho^k
>
T^2,
\end{equation}
which becomes increasingly mild as $T$ grows (Appendix~\ref{app:moderate_reweighting}).
\end{proposition}

Proposition~\ref{prop:reweight} directly shows that our reweighting remains controlled: as established by~\eqref{eq:weight_ratio_cmp}, the relative weight of a harder problem over an easier one is more moderate than that induced by the pass@$T$ objective (upper-bounded by the pass@$T$ ratio divided by $c_T(\rho)$). Such moderation is desirable, since excessively emphasizing low-success problems may introduce stronger interference across prompts and potentially degrade pass@1 performance~\cite{barakat2026pass}.

\section{Experiments}

Our experiments aim to substantiate the proposed moment-based perspective, \textbf{rather than merely pursue incremental benchmark gains}. Section~\ref{sec:overall} reports the overall performance, Section~\ref{sec:ablation} analyzes the truncation order and transformation family, and Section~\ref{sec:analysis} further examines the properties of the moment-based objective family.

\begin{table*}[t]
\centering
\setlength{\tabcolsep}{3.5pt}
\begin{tabularx}{\linewidth}{@{}l*{12}{Y}@{}}
\toprule
\multirow{2}{*}{Method}
& \multicolumn{2}{c}{MATH}
& \multicolumn{2}{c}{OlymMATH}
& \multicolumn{2}{c}{AMC23}
& \multicolumn{2}{c}{AIME24}
& \multicolumn{2}{c}{AIME25}
& \multicolumn{2}{c}{Avg.} \\
\cmidrule(lr){2-3}
\cmidrule(lr){4-5}
\cmidrule(lr){6-7}
\cmidrule(lr){8-9}
\cmidrule(lr){10-11}
\cmidrule(l){12-13}
& \textit{1.7B} & \textit{4B}
& \textit{1.7B} & \textit{4B}
& \textit{1.7B} & \textit{4B}
& \textit{1.7B} & \textit{4B}
& \textit{1.7B} & \textit{4B}
& \textit{1.7B} & \textit{4B} \\
\midrule

Base
& 53.8 & 58.6
& 21.7 & 29.9
& 29.5 & 34.8
& 4.6  & 7.7
& 2.1  & 5.0
& 22.3 & 27.2 \\

GRPO
& 69.6 & 84.0
& 34.0 & \underline{48.4}
& 46.7 & 58.1
& \textbf{12.1} & 16.5
& \underline{7.3} & 18.1
& 33.9 & 45.0 \\

Pass@K
& \underline{72.4} & 82.6
& \underline{35.5} & 47.1
& 46.0 & 60.1
& 9.2 & \underline{19.6}
& \textbf{7.5} & \underline{20.0}
& \underline{34.1} & \underline{45.9} \\

DMPO
& 71.8 & \textbf{84.8}
& 32.1 & 47.8
& 42.4 & 57.7
& 10.2 & 18.3
& 6.0 & 18.8
& 32.5 & 45.5 \\

MaxRL
& 71.4 & 82.4
& 34.6 & 45.3
& \underline{46.8} & \textbf{61.3}
& 9.5 & 15.4
& 6.6 & 12.9
& 33.8 & 43.5 \\

\textbf{MMPO (ours)}
& \textbf{73.0} & \underline{84.2}
& \textbf{35.7} & \textbf{51.6}
& \textbf{48.6} & \underline{60.6}
& \underline{10.6} & \textbf{21.3}
& 5.4 & \textbf{20.4}
& \textbf{34.7} & \textbf{47.6} \\

\bottomrule
\end{tabularx}
\caption{
    Performance on five mathematical reasoning benchmarks (\%). The best results are highlighted in \textbf{bold}, and the runner-up results are \underline{underlined}. Avg. denotes the average performance across all evaluation benchmarks.
}
\label{tab:overall_performance}
\end{table*}

\subsection{Experiment Setup}
\paragraph{Training Setup.} We conduct experiments using Qwen3-1.7B/4B-Base~\cite{yang2025qwen3} as the initial policy models. All models are trained with the \texttt{verl} framework~\cite{sheng2025hybridflow} on two NVIDIA H20 GPUs using the MATH7.5K training set~\cite{hendrycks2021measuring}. We use a training batch size of $B=16$ problems, sample $G=8$ rollouts for each problem, and set the PPO mini-batch size to 16. We use a constant learning rate of $10^{-6}$ without a warm-up phase. The maximum prompt length and response length are set to 1024 and 4096 tokens, respectively. The PPO~\cite{schulman2017proximal} clipping range is set as $\epsilon=0.2$. \texttt{Math-Verify}~\cite{kydlicek2025mathverify} is employed as the rule-based verifier to better determine the correctness of generated responses. During training, we set both the sampling temperature and the top-$p$ threshold to 1.0. The prompt is provided in Appendix~\ref{app:prompt}.

\paragraph{Evaluation Setup.} The models are evaluated on five mathematical reasoning benchmarks: AMC23, AIME24, AIME25~\cite{dekoninck2026beyond}, MATH500~\cite{hendrycks2021measuring}, and OlymMATH~\cite{sun2026challenging}. These benchmarks cover mathematical problems of varying difficulty, ranging from standard competition problems to challenging Olympiad-level reasoning tasks (Detailed information of the benchmarks are provided in Appendix~\ref{app:benchmarks}). Owing to differences in benchmark size, we report avg@1 on MATH500 and OlymMATH, and avg@16 on the remaining benchmarks. The models are evaluated every 20 training steps and we report the best average performance. During evaluation, we use a sampling temperature of 0.6 and a top-$p$ value of 0.95.

\paragraph{Method Setup.} We compare MMPO with several representative reinforcement learning objectives. GRPO~\cite{shao2024deepseekmath} and pass@$K$ training~\cite{walder2026pass} optimize objectives centered on a single moment of the failure-probability distribution, whereas DMPO~\cite{li2026beyond} promotes exploration through distribution matching. We further include MaxRL~\cite{tajwar2026maximum}, whose objective can be interpreted as optimizing the moments of a transformed random variable rather than those of $F_\theta$. For MMPO, we set the truncation order to $T=4$ and use the plug-in advantage estimator in Eq.~\eqref{eq:biased_advantage}. For pass@$K$ training, we adopt the estimator proposed in~\cite{chen2025pass} and set $K=3$, which we found to perform better than the other configurations considered. For DMPO, we follow the recommended configuration and set $\lambda=2$ and $\alpha=1/15$.

\subsection{Overall Performance}
\label{sec:overall}

\begin{figure}[t]
    \centering
    \includegraphics[width=\columnwidth]{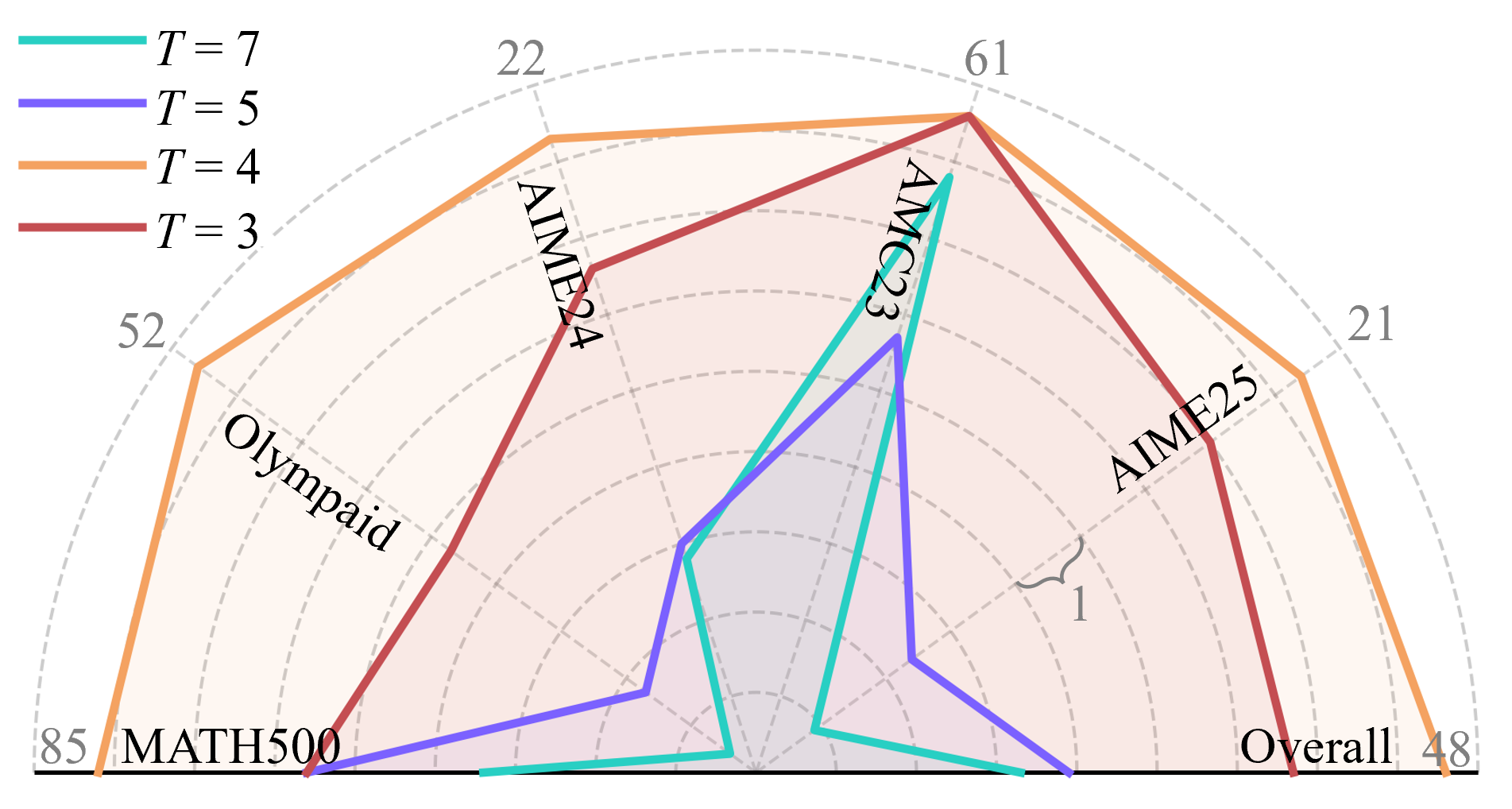}
    \caption{Ablation study of the truncation order $T$ on Qwen3-4B-Base. The y-axis shows success rate (\%).}
    \label{fig:T_ablation}
\end{figure}

As shown in Table~\ref{tab:overall_performance}, MMPO consistently achieves the best average performance across both model scales, consistently outperforming strong baselines. Compared with GRPO, MMPO improves the average score by $0.8\%$ and $2.6\%$ on Qwen3-1.7B-Base and Qwen3-4B-Base, respectively. On the 1.7B model, most alternative objectives yield only marginal or no gains over GRPO, which may be attributed to the relatively limited reasoning capacity of models at this scale and the correspondingly smaller room for improvement through objective design. Compared with Pass@$K$ training, MMPO further improves the average performance by $1.7\%$ on the 4B scale. This advantage is consistent with our Controlled Reweighting property: MMPO places greater emphasis on difficult problems while avoiding the overly aggressive reweighting induced by Pass@$K$ training. We also observe that MaxRL, the most closely related baseline, performs less favorably even when its rollout budget is increased to $G=16$, which suggests that the effectiveness of a moment objective depends jointly on the choice of transformation variable $U$ and the distribution of the training data. Consequently, the generalized moment-transformation family provide a broader design space for adapting policy objectives to different reasoning settings.

\subsection{Ablation Study}
\label{sec:ablation}

\begin{table}[t]
\centering
{
\small
\setlength{\tabcolsep}{2pt}
\begin{tabular}{lccccc}
\toprule
Transformation $U$ & MATH & Olymp. & AMC23 & AIME24 & AIME25 \\
\midrule
$U \sim \mathrm{Beta}(1,2)$ & 82.2 & 47.0 & 61.1 & 17.1 & 21.5 \\
$U \equiv 0.8$              & 83.6 & 46.5 & 58.9 & 18.3 & 17.5 \\
$U \equiv 1.0$              & 84.2 & 51.6 & 60.6 & 21.3 & 20.4 \\
\bottomrule
\end{tabular}
}
\caption{Performance of different transformation variables $U$ on Qwen3-4B-Base, with truncation order $T=4$.}
\label{tab:U_ablation}
\end{table}

We first conduct an ablation study on the truncation order $T$, which determines the number of moments of $F_\theta$ included in the MMPO objective. As shown in Figure~\ref{fig:T_ablation}, $T=4$ achieves the best overall performance, followed by $T=3$, whereas using higher truncation orders leads to a substantial performance degradation. This result reflects a practical bias--variance trade-off. With a small $T$, the objective incorporates only a limited portion of the moment profile and is therefore less capable of characterizing the distributional structure of $F_\theta$. In contrast, under finite batch size and rollout budget, higher-order moments become increasingly difficult to estimate reliably, introducing greater estimation noise and weakening the training signal. The superior performance at an intermediate order is consistent with the discussion in Section~\ref{sec:method:theory}, since a larger rollout budget implies more accurate estimation of higher-order moments and thereby enables more reliable optimization over a broader moment range.

We further examine the effect of the transformation variable $U$, which shapes the distribution targeted by the optimization objective. As shown in Table~\ref{tab:U_ablation}, different choices of $U$ lead to distinct performance. Although $U\equiv0.8$ still incorporates higher-order moments, the additional information they provide appears insufficient to offset the associated estimation noise. In comparison, $U\equiv1.0$ achieves the strongest overall performance, suggesting that uniform moment weighting is more effective in our setting.

\subsection{In-Depth Analysis}
\label{sec:analysis}

\begin{figure}[t]
    \centering
    \begin{subfigure}[t]{0.49\linewidth}
        \centering
        \includegraphics[width=\linewidth]
        {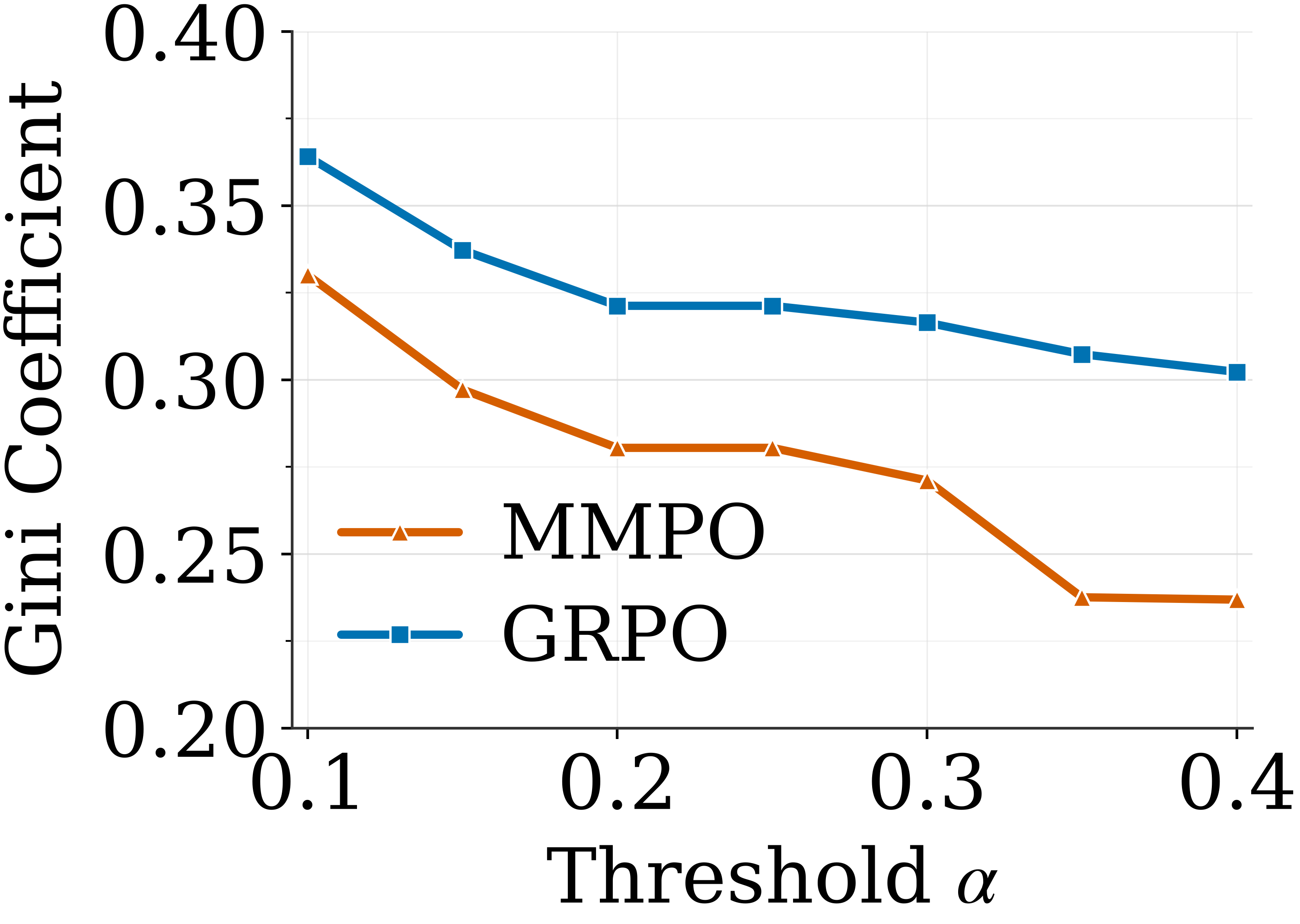}
    \end{subfigure}
    \begin{subfigure}[t]{0.49\linewidth}
        \centering
        \includegraphics[width=\linewidth]
        {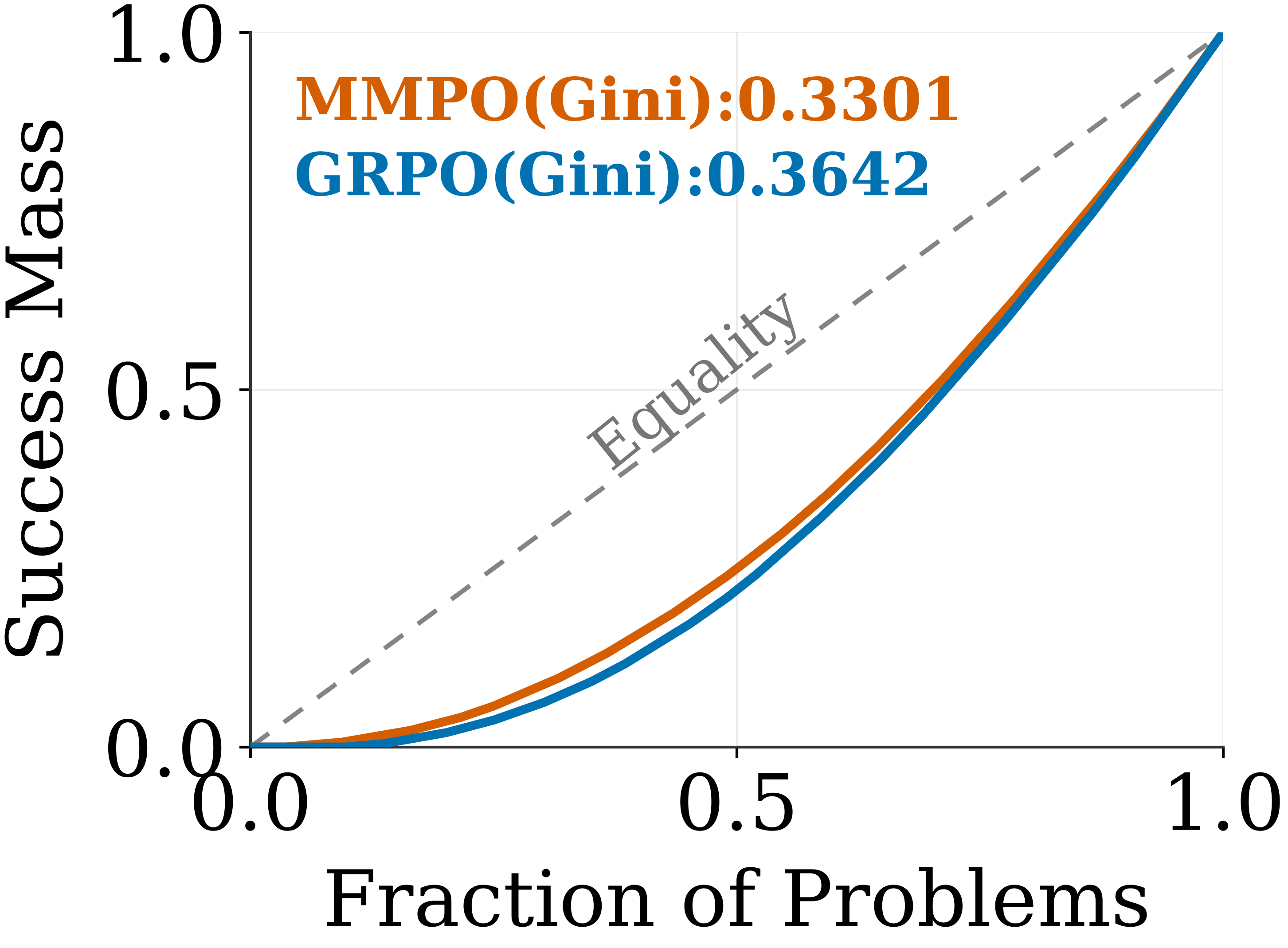}
    \end{subfigure}
    \caption{Distributional comparison between MMPO and GRPO. Left: Gini coefficients across difficulty-controlled subsets, excluding problems according to threshold $\alpha$. Right: Lorenz curves for the representative setting $\alpha=0.1$.
    }
    \label{fig:lorenz_analysis}
\end{figure}

\begin{figure}[t]
    \centering
    \includegraphics[width=\linewidth]{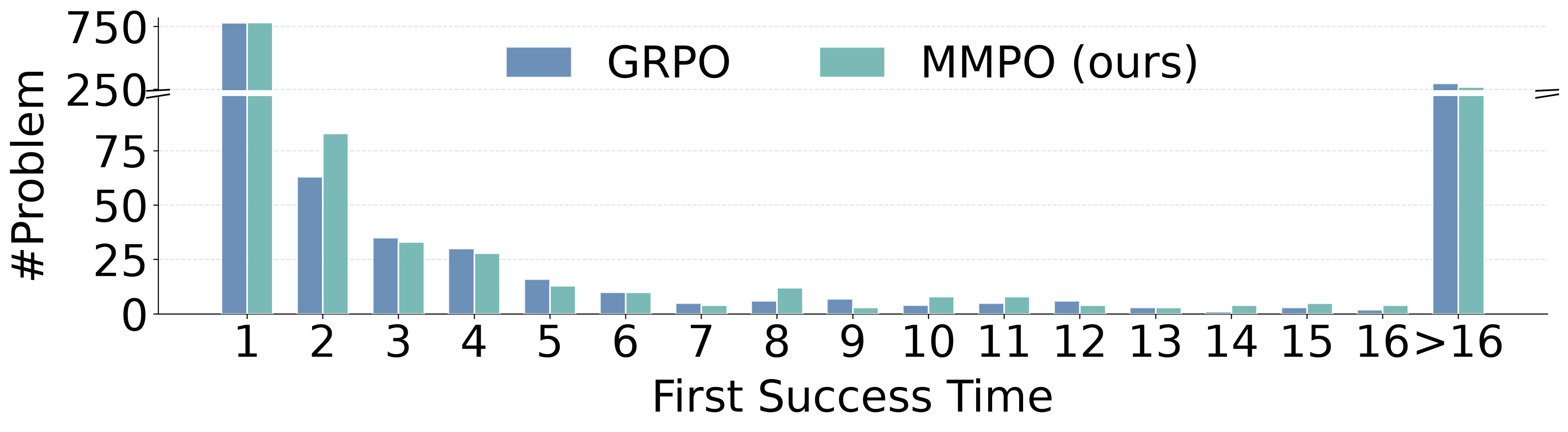}
    \caption{Comparison of the first success time distribution. The gap is less than 5 for problems solved in the first rollout, and exceeds 30 for problems unsolved after 16 rollouts.}
    \label{fig:first_success_dist}
\end{figure}

\begin{table}[t]
\centering
\setlength{\tabcolsep}{4pt}
\begin{tabular}{lcccc}
\toprule
Method & Pass@1 & Pass@4 & Pass@8 & Pass@16 \\
\midrule
GRPO        & 45.0 & 56.2 & 59.8 & 62.2 \\
MMPO (ours) & 47.6 & 59.4 & 63.4 & 66.2 \\
\bottomrule
\end{tabular}
\caption{Average Pass@$K$ performance across five mathematical reasoning benchmarks (\%).}
\label{tab:passk_results}
\end{table}

Beyond benchmark performance, we further examine how MMPO affects the distribution and dynamics of successful reasoning. Figure~\ref{fig:lorenz_analysis} presents the Lorenz curves and corresponding Gini coefficients, which characterize how evenly per-problem success probabilities are distributed (see Appendix~\ref{app:gini_lorenz} for details on the Gini coefficient and Lorenz curve). For each threshold $\alpha$, we exclude validation problems on which both methods achieve success rates below $\alpha$ or above $1-\alpha$, thereby removing instances that are uniformly too difficult or too easy, and compute the Gini coefficient over the remaining subset. Across all thresholds, MMPO consistently yields lower Gini coefficients than GRPO; under the same $\alpha$, its Lorenz curve also lies closer to the equality line, indicating a more balanced distribution of success rates across problems. This empirical pattern is consistent with the Schur-convexity property established in Theorem~\ref{theorem:schur_convex_theorem}, which shows that MMPO prefers improvements distributed across problems rather than gains concentrated on already solvable instances. The first success time distribution in Figure~\ref{fig:first_success_dist} provides a complementary perspective. The two methods solve a similar number of problems on the first rollout, suggesting that MMPO's improvement does not primarily arise from further solving easy problems. Instead, MMPO enables more problems to be solved within subsequent attempts and leaves substantially fewer problems unsolved after 16 rollouts, indicating more broadly distributed and earlier discovery of successful solutions under repeated sampling.

Table~\ref{tab:passk_results} further shows that MMPO consistently outperforms GRPO across all pass@$K$ metrics, with its advantage becoming more pronounced as $K$ increases. Since pass@$K$ metric measures the ability to discover at least one successful solution through repeated sampling, these results provide evidence that MMPO promotes more effective exploration of the solution space. We also observe that MMPO yields longer responses during training, a pattern that is consistent with more extensive reasoning trajectories and has also been reported in prior studies of exploratory behavior~\cite{guo2025deepseek, yu2026dapo, li2026beyond}.

Together, these results suggest that MMPO encourages broader exploration rather than merely reinforcing solutions that have already been discovered. More broadly, these findings highlight the effectiveness of multi-moment objectives in jointly shaping the performance, distributional balance, and exploratory behavior of reasoning policies.

\section{Conclusion}
In this paper, we introduce a moment-based perspective on policy optimization for LLM reasoning and propose Multi-Moment Policy Optimization (MMPO), which jointly optimizes multiple moments of the failure-probability distribution. MMPO admits an operational interpretation as minimizing the expected truncated time to first success, while its theoretical properties reveal a preference for balanced improvements and controlled emphasis on harder problems. We further developed a generalized moment-transformation framework that unifies a broader family of objectives. Experiments across five mathematical reasoning benchmarks and two model scales demonstrate the effectiveness of MMPO, highlighting moments as a principled foundation for designing future policy optimization objectives.

\appendix

\section{Proofs and Derivations}
\label{app:proofs}

\subsection{Unbiased Gradient Estimator}
\label{app:unbiased_estimator}

We derive an unbiased estimator of the policy-gradient direction associated with the multi-moment objective. For a problem $x$, let $s_\theta(x)=\mathbb{E}_{y\sim\pi_\theta(\cdot\mid x)} [r(x,y)]$ and $f_\theta(x)=1-s_\theta(x)$. The multi-moment objective is
\begin{equation}
\mathcal{J}_T(\theta)
=
\mathbb{E}_{x\sim\mathcal{D}}
\left[
\sum_{k=1}^{T} f_\theta(x)^k
\right],
\end{equation}
and its descent direction can be written as
\begin{equation}
-\nabla_\theta\mathcal{J}_T(\theta)
=
\mathbb{E}_{x\sim\mathcal{D}}
\left[
\sum_{k=1}^{T}
k f_\theta(x)^{k-1}
\nabla_\theta s_\theta(x)
\right].
\end{equation}
It therefore suffices to construct, for each $k\in\{1,\ldots,T\}$, an unbiased estimator of $k f_\theta(x)^{k-1}\nabla_\theta s_\theta(x)$. Fix a problem $x_i$, and let $R_{i,j}\coloneqq r(x_i,y_{i,j})$, where $\{y_{i,j}\}_{j\leq G}$ are drawn i.i.d.\ from $\pi_\theta(\cdot\mid x_i)$. \textbf{For notational simplicity}, we write $s_i\coloneqq s_\theta(x_i)$ and $f_i\coloneqq f_\theta(x_i)$. Unless otherwise specified, expectations are taken over the
joint randomness of $\{y_{i,j}\}_{j\leq G}$ conditional on $x_i$. For the $j$-th rollout, define
\begin{equation}
M_{i,-j}
=
\sum_{\ell\neq j}
\left(1-R_{i,\ell}\right),
\end{equation}
which is the number of failures among the remaining $G-1$ rollouts excluding the $j$-th rollout.

\paragraph{Unbiased estimation of $f_i^{k-1}$.} We first review the subset-averaging construction underlying the unbiased pass@$K$ estimator~\cite{walder2026pass}. For the $(k-1)$-th power of the failure probability, define
\begin{equation}
\widehat{f_{i,-j}^{k-1}}
\coloneqq
\frac{\binom{M_{i,-j}}{k-1}}
     {\binom{G-1}{k-1}},
\qquad
1\leq k\leq G,
\end{equation}
where $\binom{a}{b}$ denotes the binomial coefficient, with the convention that $\binom{a}{b}=0$ whenever $b>a$. Let $\mathcal{I}_{-j}=\{1,\ldots,G\}\setminus\{j\}$. $\widehat{f_{i,-j}^{k-1}}$ admits the equivalent $U$-statistic representation
\begin{equation}
\widehat{f_{i,-j}^{k-1}}
=
\frac{1}{\binom{G-1}{k-1}}
\sum_{\substack{
\mathcal{S}\subseteq\mathcal{I}_{-j}\\
|\mathcal{S}|=k-1
}}
\prod_{\ell\in\mathcal{S}}
\left(1-R_{i,\ell}\right).
\end{equation}
Indeed, exactly $\binom{M_{i,-j}}{k-1}$ subsets of size
$k-1$ consist entirely of failed rollouts. For every fixed
subset $\mathcal{S}\subseteq\mathcal{I}_{-j}$ satisfying
$|\mathcal{S}|=k-1$, independence of the sampled rollouts
gives

\begin{equation}
\begin{aligned}
&\mathbb{E}
\left[
\prod_{\ell\in\mathcal{S}}
\left(1-R_{i,\ell}\right)
\,\middle|\,x_i
\right]
=
\prod_{\ell\in\mathcal{S}}
\mathbb{E}
\left[
1-R_{i,\ell}
\,\middle|\,x_i
\right]
\\
&=\left(\mathbb{E}[
1-R_{i,1}
\,\middle|\,x_i]\right)^{|\mathcal{S}|}
=
f_i^{k-1}.
\end{aligned}
\end{equation}
Averaging over all subsets therefore yields
\begin{equation}
\label{app_eq:unbiased_failure_power}
\mathbb{E}
\left[
\widehat{f_{i,-j}^{k-1}}
\,\middle|\,x_i
\right]
=
f_i^{k-1}.
\end{equation}
Thus, $\widehat{f_{i,-j}^{k-1}}$ is an unbiased
estimator of $f_i^{k-1}$.

\paragraph{Leave-one-out baseline.} Define the leave-one-out estimate of the success probability $s_i$ as
\begin{equation}
\widehat{s}_{i,-j}
=
\frac{1}{G-1}\sum_{\ell\neq j}R_{i,\ell}
=
1-\frac{M_{i,-j}}{G-1},
\end{equation}
which is an unbiased estimator of $s_i$, i.e., $\mathbb{E}[\widehat{s}_{i,-j}\mid x_i]=s_i$. Both $\widehat{f_{i,-j}^{k-1}}$ and $\widehat{s}_{i,-j}$ depend only on the rollouts excluding $y_{i,j}$. Hence, conditional on $x_i$, they are independent of $y_{i,j}$. Let $g_{i,j}\coloneqq\nabla_\theta
\log\pi_\theta(y_{i,j}\mid x_i)$. The score-function
identities give $\mathbb{E}[g_{i,j}\mid x_i]=0$ and
\begin{equation}
\mathbb{E}
\left[
R_{i,j}g_{i,j}
\,\middle|\,x_i
\right]
=
\nabla_\theta s_\theta(x_i).
\end{equation}
Conditioning on the leave-one-out rollouts, we obtain
\begin{equation}
\label{app_eq:loo_baseline_identity}
\begin{aligned}
&\mathbb{E}_{y_{i,j}\sim\pi_\theta(\cdot\mid x_i)}
\left[
\left(
R_{i,j}-\widehat{s}_{i,-j}
\right)
g_{i,j}
\,\middle|\,
x_i,\{y_{i,\ell}\}_{\ell\neq j}
\right]
\\
&\quad=
\mathbb{E}
\left[
R_{i,j}g_{i,j}
\,\middle|\,x_i
\right]
-
\widehat{s}_{i,-j}
\mathbb{E}
\left[
g_{i,j}
\,\middle|\,x_i
\right]
\\
&\quad=
\nabla_\theta s_\theta(x_i).
\end{aligned}
\end{equation}
Therefore, the leave-one-out baseline does not change the
expected score-weighted gradient.

\paragraph{Unbiased estimator for each moment.} For the $k$-th moment, define the leave-one-out advantage coefficient
\begin{equation}
\widehat{A}_{i,j}^{(k)}
=
k
\widehat{f_{i,-j}^{k-1}}
\left(
R_{i,j}-\widehat{s}_{i,-j}
\right).
\end{equation}
Equivalently,
\begin{equation}
\label{app_eq:unbiased_single_moment_advantage}
\widehat{A}_{i,j}^{(k)}
=
k
\frac{\binom{M_{i,-j}}{k-1}}
     {\binom{G-1}{k-1}}
\left(
R_{i,j}
-
1
+
\frac{M_{i,-j}}{G-1}
\right).
\end{equation}
Using the tower property of conditional expectation gives
\begin{equation}
\label{app_eq:unbiased_single_moment_gradient}
\begin{aligned}
&\mathbb{E}
\left[
\widehat{A}_{i,j}^{(k)}g_{i,j}
\,\middle|\,x_i
\right]\xlongequal{\text{tower property}}
k\,
\mathbb{E}_{y_{i,\ell}\sim\pi_\theta(\cdot\mid x_i),\,\ell\neq j}\Bigg[
\widehat{f_{i,-j}^{k-1}}
\,
\\
&\mathbb{E}_{y_{i,j}\sim\pi_\theta(\cdot\mid x_i)}
\left[
\left(
R_{i,j}-\widehat{s}_{i,-j}
\right)g_{i,j}
\,\middle|\,
x_i,\{y_{i,\ell}\}_{\ell\neq j}
\right]
\,\Bigg|\,
x_i
\Bigg]
\\
&\quad\xlongequal{\text{Eq.~\eqref{app_eq:loo_baseline_identity}}}k\,
\mathbb{E}_{y_{i,\ell}\sim\pi_\theta(\cdot\mid x_i),\,\ell\neq j}\Bigg[
\widehat{f_{i,-j}^{k-1}}
\,\nabla_\theta s_\theta(x_i)
\,\Bigg|\,
x_i
\Bigg]
\\
&\quad=k\,
\mathbb{E}\Bigg[
\widehat{f_{i,-j}^{k-1}}
\,\nabla_\theta s_\theta(x_i)
\,\Bigg|\,
x_i
\Bigg]
\\
&\quad=
k\,
\mathbb{E}
\left[
\widehat{f_{i,-j}^{k-1}}
\,\middle|\,x_i
\right]
\nabla_\theta s_\theta(x_i)
\\
&\quad\xlongequal{\text{Eq.~\eqref{app_eq:unbiased_failure_power}}}
k f_i^{k-1}\nabla_\theta s_\theta(x_i).
\end{aligned}
\end{equation}
Hence, $\widehat{A}_{i,j}^{(k)}
\nabla_\theta\log\pi_\theta(y_{i,j}\mid x_i)$ is an unbiased estimator of the policy-gradient contribution of the $k$-th moment.

\paragraph{Unbiased multi-moment estimator.} Using Eq.~\eqref{app_eq:unbiased_single_moment_advantage}, summing over the first $T$ moments gives
\begin{equation}
\label{app_eq:unbiased_multi_moment_advantage}
\widehat{A}_{i,j}^{\mathrm{unb}}
=
\left[
\sum_{k=1}^{T}
k
\frac{\binom{M_{i,-j}}{k-1}}
     {\binom{G-1}{k-1}}
\right]
\left(
R_{i,j}
-
1
+
\frac{M_{i,-j}}{G-1}
\right).
\end{equation}
Its expected score-weighted value satisfies

\begin{equation}
\begin{aligned}
&\mathbb{E}
\left[
\widehat{A}_{i,j}^{\mathrm{unb}}
\nabla_\theta
\log\pi_\theta(y_{i,j}\mid x_i)
\,\middle|\,x_i
\right]
\\
&\quad\xlongequal{\text{Eq.~\eqref{app_eq:unbiased_single_moment_gradient}}}
\sum_{k=1}^{T}
k f_\theta^{k-1}
\nabla_\theta s_\theta(x_i)
\\
&\quad\xlongequal{\text{Eq.~\eqref{eq:mmpo_weight}}}
w_{T,\theta}(x_i)
\nabla_\theta s_\theta(x_i).
\end{aligned}
\end{equation}
Consequently, whenever $T\leq G$,
\begin{equation}
\frac{1}{BG}
\sum_{i=1}^{B}
\sum_{j=1}^{G}
\widehat{A}_{i,j}^{\mathrm{unb}}
\nabla_\theta
\log\pi_\theta(y_{i,j}\mid x_i)
\end{equation}
is an unbiased Monte Carlo estimator of the gradient in~\eqref{eq:original_mmpo_gradient}.

\paragraph{Simplified form when $T=G$.} We next simplify the estimator for $T=G$. For
$0\leq m\leq G-1$, define
\begin{equation}
S_G(m)
\coloneqq
\sum_{k=1}^{G}
k
\frac{\binom{m}{k-1}}
     {\binom{G-1}{k-1}}
=\sum_{t=0}^{m}
(t+1)
\frac{\binom{m}{t}}
     {\binom{G-1}{t}},
\end{equation}
For $0\leq t\leq G-1$, the beta-integral identity gives
\begin{equation}
\frac{1}{\binom{G-1}{t}}
=
G\int_{0}^{1}
u^t(1-u)^{G-1-t}\,du,
\end{equation}
we obtain
\begin{equation}
\begin{aligned}
S_G(m)
&=
G\int_{0}^{1}
(1-u)^{G-1-m}
\\
&\qquad\cdot
\left[
\sum_{t=0}^{m}
(t+1)\binom{m}{t}
u^t(1-u)^{m-t}
\right]du.
\end{aligned}
\end{equation}
The standard binomial identities imply

\begin{equation}
\sum_{t=0}^{m}
(t+1)\binom{m}{t}
u^t(1-u)^{m-t}
=
1+mu.
\end{equation}
Therefore,
\begin{equation}
S_G(m)
=
G\int_{0}^{1}
(1-u)^{G-1-m}(1+mu)\,du.
\end{equation}
Evaluating the integral yields

\begin{equation}
\begin{aligned}
S_G(m)
&=
G\left[
\frac{1}{G-m}
+
\frac{m}{(G-m)(G-m+1)}
\right]
\\
&=
\frac{G(G+1)}
     {(G-m)(G-m+1)}.
\end{aligned}
\end{equation}
Applying this result with $m=M_{i,-j}$ allows us to simplify the summation term in Eq.~\eqref{app_eq:unbiased_multi_moment_advantage}. We next consider the two cases corresponding to
$R_{i,j}=1$ and $R_{i,j}=0$, respectively.

Let $N_i=\sum_{j=1}^{G}R_{i,j}$ denote the total number of successful rollouts for problem $x_i$. When $R_{i,j}=1$, we have
$M_{i,-j}=G-N_i$ and
$R_{i,j}
-
1
+
\frac{M_{i,-j}}{G-1}=\frac{G-N_i}{G-1}$.
Thus,
\begin{equation}
\begin{aligned}
\widehat{A}_{i,j}^{\mathrm{unb}}
&=
S_G(G-N_i)\frac{G-N_i}{G-1}\\
&=
\frac{G(G+1)(G-N_i)}
     {(G-1)N_i(N_i+1)}.
\end{aligned}
\end{equation}
When $R_{i,j}=0$, we have
$M_{i,-j}=G-N_i-1$ and
$R_{i,j}
-
1
+
\frac{M_{i,-j}}{G-1}=-\frac{N_i}{G-1}$.
Consequently,

\begin{equation}
\begin{aligned}
\widehat{A}_{i,j}^{\mathrm{unb}}
&=
-S_G(G-N_i-1)\frac{N_i}{G-1}
\\
&=
-\frac{G(G+1)N_i}
       {(G-1)(N_i+1)(N_i+2)}.
\end{aligned}
\end{equation}
Combining the two cases gives the simplified form Eq.~\eqref{eq:mmpo_simple_case}.

\paragraph{Bias-variance trade-off.} The derivation above establishes an unbiased estimator for the multi-moment gradient. Nevertheless, unbiasedness alone does not guarantee low finite-sample estimation error or stable optimization, motivating the deliberate bias-variance trade-offs commonly adopted in policy-gradient methods~\cite{schulman2015high}. We therefore use the generally biased plug-in estimator in Eq.~\eqref{eq:biased_advantage} throughout the main experiments. With a limited rollout budget and batch size, this choice accepts finite-sample bias in favor of reduced estimation noise and improved gradient stability. It thus represents a deliberate bias--variance trade-off in the practically relevant limited-sample regime, rather than a modification of the underlying multi-moment objective. The resulting improvements further demonstrate that the proposed multi-moment signal remains effective under practical sampling constraints. Meanwhile, the unbiased estimator remains directly applicable and, for fixed $T$, is expected to become increasingly stable as budgets grow. We further compare the two estimators in Appendix~\ref{app:bias_variance_tradeoff}.

\subsection{Schur-Convexity of Moment Objectives}
\label{app:schur_convexity}

For clarity, we first specify the majorization notation used in our paper. We write $\mathbf{f}\succeq\mathbf{g}$ if $\mathbf{f}$ majorizes $\mathbf{g}$, and write $\mathbf{f}\succ\mathbf{g}$ if, in addition, $\mathbf{g}$ is not a permutation of $\mathbf{f}$. Define the scalar
\begin{equation}
\phi_{U,T}(z)
\coloneqq
\sum_{k=1}^{T}\mathbb{E}[U^k]z^k,
\qquad z\in[0,1].
\end{equation}
Then, the objective can be written as
\begin{equation}
J_{U,T}(\mathbf{f})
=
\frac{1}{n}\sum_{i=1}^{n}\phi_{U,T}(f_i).
\end{equation}
We first show that $\phi_{U,T}$ is strictly convex on $[0,1]$. Since $T\geq 2$, its second derivative is
\begin{equation}
\begin{aligned}
\phi_{U,T}''(z)
&=
\sum_{k=2}^{T}
k(k-1)\mathbb{E}[U^k]z^{k-2} \\
&=
2\mathbb{E}[U^2]
+
\sum_{k=3}^{T}
k(k-1)\mathbb{E}[U^k]z^{k-2}.
\end{aligned}
\end{equation}
Since $U\in[0,1]$ and $\Pr(U>0)>0$, we have $\mathbb{E}[U^2]>0$. Moreover, all the remaining terms are nonnegative for $z\in[0,1]$. Therefore,
\begin{equation}
\phi_{U,T}''(z)
\geq
2\mathbb{E}[U^2]
>
0,
\qquad z\in[0,1].
\end{equation}
Thus, $\phi_{U,T}$ is strictly convex on $[0,1]$. Now suppose that $\mathbf{f}\succeq\mathbf{g}$. By Karamata's inequality~\cite{marshall2011inequalities}, the convexity of $\phi_{U,T}$ gives
\begin{equation}
\sum_{i=1}^{n}\phi_{U,T}(f_i)
\geq
\sum_{i=1}^{n}\phi_{U,T}(g_i).
\end{equation}
Since $\phi_{U,T}$ is strictly convex, equality holds only when $\mathbf{g}$ is a permutation of $\mathbf{f}$~\cite{marshall2011inequalities}. Therefore, when $\mathbf{f}\succ\mathbf{g}$, we have
\begin{equation}
\sum_{i=1}^{n}\phi_{U,T}(f_i)
>
\sum_{i=1}^{n}\phi_{U,T}(g_i).
\end{equation}
Dividing both sides by $n$ yields
\begin{equation}
J_{U,T}(\mathbf{f})
>
J_{U,T}(\mathbf{g}).
\end{equation}
Therefore, $J_{U,T}$ is strictly Schur-convex. It is worth noting that this conclusion does not hold when $T=1$. In this case, $\phi_{U,1}(z)=\mathbb{E}[U]z$ is linear rather than strictly convex. Since majorization requires $\sum_{i=1}^{n}f_i=\sum_{i=1}^{n}g_i$, we have
\begin{equation}
J_{U,1}(\mathbf{f})
=
\frac{\mathbb{E}[U]}{n}\sum_{i=1}^{n}f_i
=
\frac{\mathbb{E}[U]}{n}\sum_{i=1}^{n}g_i
=
J_{U,1}(\mathbf{g}).
\end{equation}
Thus, a first-moment objective cannot distinguish failure-probability vectors with the same mean but different levels of dispersion. This highlights an inherent limitation of first-moment optimization: it captures average performance but is insensitive to how performance is distributed~\cite{bellemare2017distributional}.

\subsection{Moderate Reweighting Property}
\label{app:moderate_reweighting}

For brevity, write
$a=f_\theta(x_{\mathrm{easy}})$ and $t=f_\theta(x_{\mathrm{hard}})$, so that $t=\rho a$. Since failure probabilities lie in $(0,1)$, we have $a,t\in(0,1)$. Recall that the problem-level weights induced by the multi-moment and pass@$T$
objectives are, respectively,
\begin{equation}
w_{T,\theta}(x)
=
\sum_{k=1}^{T} k f_\theta(x)^{k-1},
\,\,
W_{T,\theta}(x)
=
T f_\theta(x)^{T-1}.
\end{equation}
We first divide the relative weight induced by pass@$T$ by that induced
by the multi-moment objective:
\begin{equation}
\begin{aligned}
&\frac{
    W_{T,\theta}(x_{\mathrm{hard}})
    /
    W_{T,\theta}(x_{\mathrm{easy}})
}{
    w_{T,\theta}(x_{\mathrm{hard}})
    /
    w_{T,\theta}(x_{\mathrm{easy}})
}\\
&=
\rho^{T-1}
\frac{\sum_{k=1}^{T}k a^{k-1}}
     {\sum_{k=1}^{T}k(\rho a)^{k-1}} \\
&=
\frac{
    \sum_{k=1}^{T}
    k\rho^{T-k}t^{k-1}
}{
    \sum_{k=1}^{T}
    kt^{k-1}
}.
\end{aligned}
\end{equation}
Define
\begin{equation}
\label{app_eq:defination_pbq}
p_k
\coloneqq
\frac{2k}{T(T+1)},
\qquad
b_k
\coloneqq
\rho^{T-k},
\qquad
q_k
\coloneqq
t^{k-1},
\end{equation}
where the sequence $\{b_k\}_{k=1}^{T}$ and $\{q_k\}_{k=1}^{T}$ are decreasing in $k$. Noting that $\sum_{k=1}^{T}p_k=1$, direct expansion (or, equivalently, the weighted Chebyshev sum inequality) gives
\begin{equation}
\begin{aligned}
&\sum_{k=1}^{T}p_kb_kq_k
-
\left(\sum_{k=1}^{T}p_kb_k\right)
\left(\sum_{k=1}^{T}p_kq_k\right) \\
&\qquad =
\frac{1}{2}
\sum_{i=1}^{T}\sum_{j=1}^{T}
p_ip_j(b_i-b_j)(q_i-q_j)
\geq 0.
\end{aligned}
\end{equation}
Dividing by $\sum_{k=1}^{T}p_kq_k>0$ and substituting Eq.~\eqref{app_eq:defination_pbq} gives
\begin{equation}
\frac{
    \sum_{k=1}^{T}
    k\rho^{T-k}t^{k-1}
}{
    \sum_{k=1}^{T}
    kt^{k-1}
}
\geq
\frac{
    2\sum_{k=1}^{T}k\rho^{T-k}
}{
    T(T+1)
}\geq 1.
\end{equation}
The middle term is exactly $c_T(\rho)$. Consequently,
\begin{equation}
\frac{
    w_{T,\theta}(x_{\mathrm{hard}})
}{
    w_{T,\theta}(x_{\mathrm{easy}})
}
\leq
\frac{1}{c_T(\rho)}
\frac{
    W_{T,\theta}(x_{\mathrm{hard}})
}{
    W_{T,\theta}(x_{\mathrm{easy}})
}.
\end{equation}
To characterize when $c_T(\rho)>2$, consider
\begin{equation}
\sum_{k=1}^{T}k\rho^{T-k}
=
T+
\sum_{k=1}^{T-1}(T-k)\rho^k.
\end{equation}
It follows that
\begin{equation}
\begin{aligned}
c_T(\rho)>2
&\iff
T+\sum_{k=1}^{T-1}(T-k)\rho^k
>
T(T+1) \\
&\iff
\sum_{k=1}^{T-1}(T-k)\rho^k
>
T^2.
\end{aligned}
\end{equation}
To show that this condition becomes easier to satisfy as $T$ increases, in the sense that the minimum required value of $\rho$ decreases, we define
\begin{equation}
g_T(\rho)
\coloneqq
\frac{1}{T^2}
\sum_{k=1}^{T-1}(T-k)\rho^k .
\end{equation}
For any fixed $\rho>1$,
\begin{equation}
g_{T+1}(\rho)-g_T(\rho)
=
\frac{
\displaystyle\sum_{k=1}^{T}
\bigl((2T+1)k-T(T+1)\bigr)\rho^k
}{
T^2(T+1)^2
}
>0,
\end{equation}
where the inequality follows from
\begin{equation}
\begin{aligned}
\frac{\sum_{k=1}^{T}k\rho^k}
     {\sum_{k=1}^{T}\rho^k}
&>
\frac{T+1}{2}\text{(by the Chebyshev inequality)}\\
&>
\frac{T(T+1)}{2T+1}.
\end{aligned}
\end{equation}
Therefore, $g_T(\rho)$ is increasing in $T$. Since it is also
strictly increasing in $\rho$, the minimum value of $\rho$ satisfying
$g_T(\rho)>1$ decreases as $T$ increases. For example, for $T=3,4,5$, the corresponding minimum values are
\begin{equation}
\rho_3^\star\approx 2.1623,\qquad
\rho_4^\star\approx 1.7112,\qquad
\rho_5^\star\approx 1.5109.
\end{equation}
Therefore, our objective not only upweights harder problems, but also exhibits a moderate reweighting behavior, in the sense that it avoids the overly aggressive emphasis induced by the corresponding pass@$T$ objective.

\section{Additional Experimental Details}
\label{app:experiments}

\subsection{A Toy Example on Moments}
\label{app:toy_example}

Consider two policies whose per-problem failure probabilities follow
the continuous distributions
\begin{equation}
F_{\theta_1} \sim \operatorname{Unif}(0.3,0.5),
\qquad
F_{\theta_2} \sim \operatorname{Beta}\!\left(\frac{1}{2},1\right).
\end{equation}
The second distribution places substantially more probability near
zero, indicating that $\theta_2$ achieves low failure probabilities on a larger fraction of problems. Their first moments are
\begin{equation}
\mathbb{E}[F_{\theta_1}]
=
\frac{2}{5}
>
\frac{1}{3}
=
\mathbb{E}[F_{\theta_2}].
\end{equation}
Thus, the first moment alone favors $\theta_2$. In contrast, their
fourth moments satisfy
\begin{equation}
\mathbb{E}\!\left[F_{\theta_1}^{4}\right]
=
\frac{1441}{50000}
<
\frac{1}{9}
=
\mathbb{E}\!\left[F_{\theta_2}^{4}\right].
\end{equation}
Therefore, the fourth moment favors $\theta_1$. Intuitively, although
$\theta_2$ performs better on average and assigns more probability to
near-zero failure rates, it also has a heavier upper tail corresponding
to a subset of particularly difficult problems, which is characterized by higher-order moments. In comparison,
$\theta_1$ distributes its failure probabilities more evenly.

When jointly considering the first four moments, the MMPO objective
favors $\theta_1$, since
\begin{equation}
\sum_{k=1}^{4}\mathbb{E}\!\left[F_{\theta_1}^{k}\right]
=
\frac{99023}{150000}
<
\frac{248}{315}
=
\sum_{k=1}^{4}\mathbb{E}\!\left[F_{\theta_2}^{k}\right].
\end{equation}
Thus, MMPO may prefer a policy with a slightly worse average failure probability but substantially smaller higher-order moments, reflecting its greater emphasis on reducing failures on particularly difficult problems. In our experiments, MMPO consistently outperforms first-moment methods, suggesting that explicitly accounting for higher-order moments can ultimately reduce the first moment of the failure probability as well.

\subsection{Prompt Template}
\label{app:prompt}

\begin{figure}[t]
    \centering
    \includegraphics[width=\columnwidth]{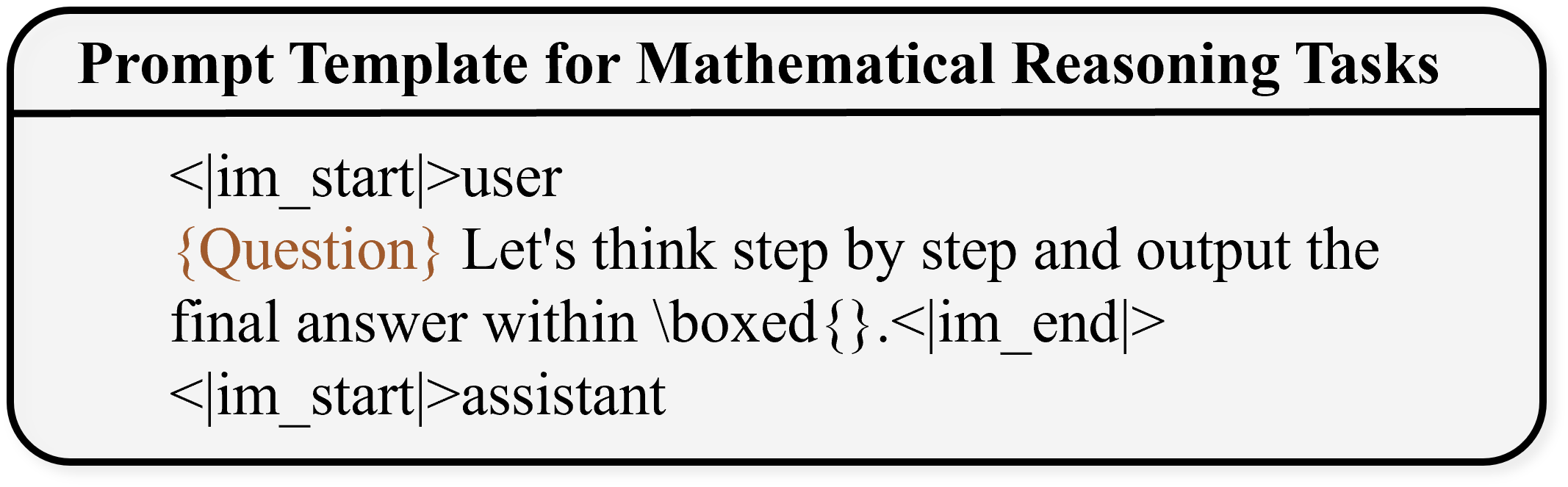}
    \caption{Prompt template.}
    \label{app_fig:prompt}
\end{figure}

Figure~\ref{app_fig:prompt} presents the prompt template used in our experiments. We construct this template using Python-style string formatting, where placeholders enclosed in curly braces are dynamically populated at runtime.

\subsection{Benchmark Details}
\label{app:benchmarks}

We focus on mathematical reasoning, a core capability of LLMs. Below, we provide details of the five mathematical reasoning benchmarks used in our experiments.
\begin{itemize}
    \item \textbf{MATH and MATH500~\cite{hendrycks2021measuring}.}
    The MATH dataset contains $12{,}500$ competition-level mathematics
    problems, divided into $7{,}500$ training problems and $5{,}000$
    test problems. We use all $7{,}500$ problems in the training split
    for reinforcement learning and evaluate on MATH500, a representative
    subset of $500$ problems drawn from the original test split.
    Therefore, the training and evaluation problems of MATH do not overlap.

    \item \textbf{OlymMATH~\cite{sun2026challenging}.}
    OlymMATH contains $200$ manually curated Olympiad-level problems. The problems
    span four major mathematical domains and are divided into easy and hard
    subsets. We use its English version for evaluation.

    \item \textbf{AMC23~\cite{dekoninck2026beyond}.}
    AMC23 consists of $40$ problems selected from the 2023 AMC 12A and
    AMC 12B competitions. It evaluates high-school-level competition
    mathematics across topics such as algebra, geometry, number theory,
    and combinatorics.

    \item \textbf{AIME24~\cite{dekoninck2026beyond}.}
    AIME24 contains $30$ problems from the 2024 AIME I and AIME II
    competitions, with $15$ problems from each examination. These problems
    require advanced high-school mathematical reasoning and have integer
    answers between $0$ and $999$.

    \item \textbf{AIME25~\cite{dekoninck2026beyond}.}
    AIME25 similarly contains $30$ problems from the 2025 AIME I and
    AIME II competitions. It provides a recent and challenging evaluation
    of mathematical reasoning, covering algebra, geometry, number theory,
    and combinatorics.
\end{itemize}
Except for the MATH7.5K training split, we combine the remaining five benchmarks into a unified validation set. Following standard evaluation practice, we evaluate each method at regular training intervals and report the checkpoint achieving the highest accuracy on this unified set. The same evaluation frequency and checkpoint-selection criterion are applied to all methods to ensure a fair comparison.

\subsection{Gini Coefficient and Lorenz Curve}
\label{app:gini_lorenz}

The Gini coefficient measures the dispersion of per-problem success probabilities. Under a common mean, a lower Gini coefficient indicates
that performance is distributed more evenly across problems, rather than concentrated on a small subset. The corresponding Lorenz curve
plots the cumulative fraction of success mass against the cumulative fraction of problems, after sorting problems by their success
probabilities. A curve closer to the $45^\circ$ equality line therefore indicates a more balanced accuracy distribution.

To ensure a controlled comparison, we independently sample $16$ responses for every problem in the unified validation set. For method $m\in\{\mathrm{MMPO},\mathrm{GRPO}\}$, the empirical success probability of problem $i$ is $
\widehat{s}_{i}^{(m)}
=
\frac{1}{16}\sum_{j=1}^{16} r_{i,j}^{(m)}.
$ For each threshold $\alpha$, we remove only problems that are commonly too difficult for both methods or commonly too easy for both methods. Specifically, the retained problem set is
\begin{equation}
\mathcal{I}_{\alpha}
=
\left\{
i:
\max_{m}
\widehat{s}_{i}^{(m)}
\geq \alpha,
\quad
\min_{m}
\widehat{s}_{i}^{(m)}
\leq 1-\alpha
\right\}.
\end{equation}
This common filtering allows us to compare their accuracy dispersion over the same set of moderately difficult problems. Let $n_{\alpha}=|\mathcal{I}_{\alpha}|$ and define the mean accuracy of
method $m$ on this subset as $
\overline{s}_{\alpha}^{(m)}
=
\frac{1}{n_{\alpha}}
\sum_{i\in\mathcal{I}_{\alpha}}
\widehat{s}_{i}^{(m)}
$. To isolate distributional differences from differences in average
accuracy, we normalize each success probabilities to a common mean:
\begin{equation}
z_{i}^{(m)}
=
\frac{\widehat{s}_{i}^{(m)}}
{\overline{s}_{\alpha}^{(m)}},
\qquad
\frac{1}{n_{\alpha}}
\sum_{i\in\mathcal{I}_{\alpha}}z_i^{(m)}=1.
\end{equation}
The Gini coefficient is then computed as
\begin{equation}
\text{Gini}_{\alpha}^{(m)}
=
\frac{1}{2n_{\alpha}^{2}}
\sum_{i\in\mathcal{I}_{\alpha}}
\sum_{j\in\mathcal{I}_{\alpha}}
\left|z_i^{(m)}-z_j^{(m)}\right|.
\end{equation}
For the Lorenz curve, let
$z_{(1)}^{(m)}\leq\cdots\leq z_{(n_{\alpha})}^{(m)}$ denote the sorted
normalized probabilities of $\mathcal{I}_{\alpha}$. Its value at $q/n_{\alpha}$ is
\begin{equation}
\text{Lorenz}_{\alpha}^{(m)}
\left(\frac{q}{n_{\alpha}}\right)
=
\frac{
\sum_{i=1}^{q}z_{(i)}^{(m)}
}{
\sum_{i=1}^{n_{\alpha}}z_{(i)}^{(m)}
},
\qquad
q=0,\ldots,n_{\alpha}.
\end{equation}

As shown in Figure~\ref{fig:lorenz_analysis}, MMPO achieves a lower
Gini coefficient for every evaluated values of $\alpha$. Its Lorenz
curve is also consistently closer to the equality line than that of GRPO~\cite{shao2024deepseekmath}, indicating that MMPO distributes its improvements more evenly across problems.

\subsection{Biased and Unbiased Estimators}
\label{app:bias_variance_tradeoff}

\begin{table}[t]
\centering
{
\small
\setlength{\tabcolsep}{2pt}
\begin{tabular}{lcccccc}
\toprule
Method & MATH & Olymp. & AMC23 & AIME24 & AIME25 & Avg.\\
\midrule
GRPO & 81.8 & \textbf{51.7} & 64.0 & 19.3 & 18.9 & 47.1 \\
Unb, $T=4$ & 84.8 & 49.7 & 65.3 & \underline{24.0} & 17.5 & 48.3 \\
Unb, $T=3$ & \underline{85.6} & \underline{51.2} & \textbf{66.3} & 23.1 & \underline{19.8} & \underline{49.2} \\
Bias & \textbf{85.8} & 50.1 & \underline{66.1} & \textbf{24.6} & \textbf{21.1} & \textbf{49.5} \\
\bottomrule
\end{tabular}
}
\caption{Comparison of the biased and unbiased MMPO estimators under a larger computational budget. The best results are highlighted in \textbf{bold}, and the runner-up results are \underline{underlined}. Avg. denotes the average performance. The last three rows correspond to MMPO using the unbiased estimator with $T=4$, the unbiased estimator with $T=3$, and the biased estimator with $T=4$, respectively.}
\label{app_tab:estimator_comparison}
\end{table}

To complement the main experiments, we further compare the two MMPO estimators under a larger computational budget. MMPO(Unb) uses the unbiased estimator in Eq.~\eqref{eq:unbiased_advantage}, whereas MMPO(Bias) uses the plug-in estimator in Eq.~\eqref{eq:biased_advantage}. We double both the batch size $B$ and the PPO mini-batch size to $32$, and increase the maximum response length to $8192$. As shown in Table~\ref{app_tab:estimator_comparison}, MMPO(Bias) still achieves the best overall performance, while MMPO(Unb, $T=4$) also obtains competitive results. In particular, compared with the main setting where MMPO(Unb, $T=4$) underperformed GRPO by $1.5\%$ on average, it now surpasses GRPO by $1.2\%$ points. This narrowed gap suggests that increasing the sampling and optimization budgets improves the stability of higher-order moment estimation, consistent with the discussion in Section~\ref{sec:method:theory}. We also observe that reducing $T$ from $4$ to $3$ improves the performance of the unbiased estimator, whereas the opposite trend is observed for the biased estimator in our ablation study. This contrast further highlights the importance of jointly considering the truncation order $T$ and the estimator's bias-variance trade-off.

\bibliography{aaai2027}


\end{document}